\documentclass[10pt,twocolumn,letterpaper]{article}

\usepackage{authblk}
\usepackage{cvpr}              % To produce the CAMERA-READY version
\usepackage{arydshln}
\usepackage{multirow}
\definecolor{cvprblue}{rgb}{0.21,0.49,0.74}
\usepackage[pagebackref,breaklinks,colorlinks,allcolors=cvprblue]{hyperref}

\def\paperID{11251} % *** Enter the Paper ID here
\def\confName{CVPR}
\def\confYear{2026}

\title{\textcolor[RGB]{166,35,35}{FUSER}:  \textcolor[RGB]{166,35,35}{F}eed-Forward M\textcolor[RGB]{166,35,35}{U}ltiview 3D Registration Transformer and  \textcolor[RGB]{166,35,35}{SE}(3)$^N$ Diffusion \textcolor[RGB]{166,35,35}{R}efinement}

\author{Haobo Jiang$^{1}$, Jin Xie$^{3}$, Jian Yang$^{3}$, Liang Yu$^{2}$, Jianmin Zheng$^{*1}$}
\affil{{$^1$Nanyang Technological University, $^2$Alibaba Group, $^3$Nanjing University}}
\affil{\tt\small \{haobo.jiang, ASJMZheng\}@ntu.edu.sg, \{csjyang, csjxie\}@nju.edu.cn, \tt\small liangyu.yl@alibaba-inc.com
}

\begin{document}
\maketitle
\let\thefootnote\relax\footnotetext{$^*$Corresponding author}
\begin{abstract}
Registration of multiview point clouds conventionally relies on extensive pairwise matching to build a pose graph for global synchronization, which is computationally expensive and inherently ill-posed without holistic geometric constraints.
This paper proposes FUSER, the first feed-forward multiview registration transformer that jointly processes all scans in a unified, compact latent space to directly predict global poses without any pairwise estimation.
To maintain tractability, FUSER encodes each scan into low-resolution superpoint features via a sparse 3D CNN that preserves absolute translation cues, and performs efficient intra- and inter-scan reasoning through a Geometric Alternating Attention module. 
Particularly, we transfer 2D attention priors from off-the-shelf foundation models to enhance 3D feature interaction and geometric consistency.
Building upon FUSER, we further introduce FUSER-DF, an SE(3)$^N$ diffusion refinement framework to correct  FUSER's estimates via denoising in the joint SE(3)$^N$ space. 
FUSER acts as a surrogate multiview registration model to construct the denoiser, and a prior-conditioned SE(3)$^N$ variational lower bound is derived for denoising supervision.  
Extensive experiments on 3DMatch, ScanNet and ArkitScenes demonstrate that our approach achieves the superior registration accuracy and outstanding computational efficiency. Code is available at \href{https://github.com/Jiang-HB/FUSER}{https://github.com/Jiang-HB/FUSER}.
\end{abstract}    
\section{Introduction}
\label{sec:intro}
Multiview point cloud registration aims to estimate the global rigid poses of a set of unordered, partially overlapping point-cloud scans, thereby aligning them into a common coordinate system. It is a fundamental yet challenging problem, playing an important role in extensive downstream applications, such as  3D scene reconstruction~\cite{newcombe2011kinectfusion,dai2017bundlefusion}, AR/VR~\cite{azuma1997survey,kim2022benchmark,zhu2024spgroup3d,Zhu_2025_CVPR}, and embodied AI~\cite{savva2019habitat,szot2021habitat,liu2025diff9d}.

\begin{figure}[t]
	\centering
	\includegraphics[width=\columnwidth]{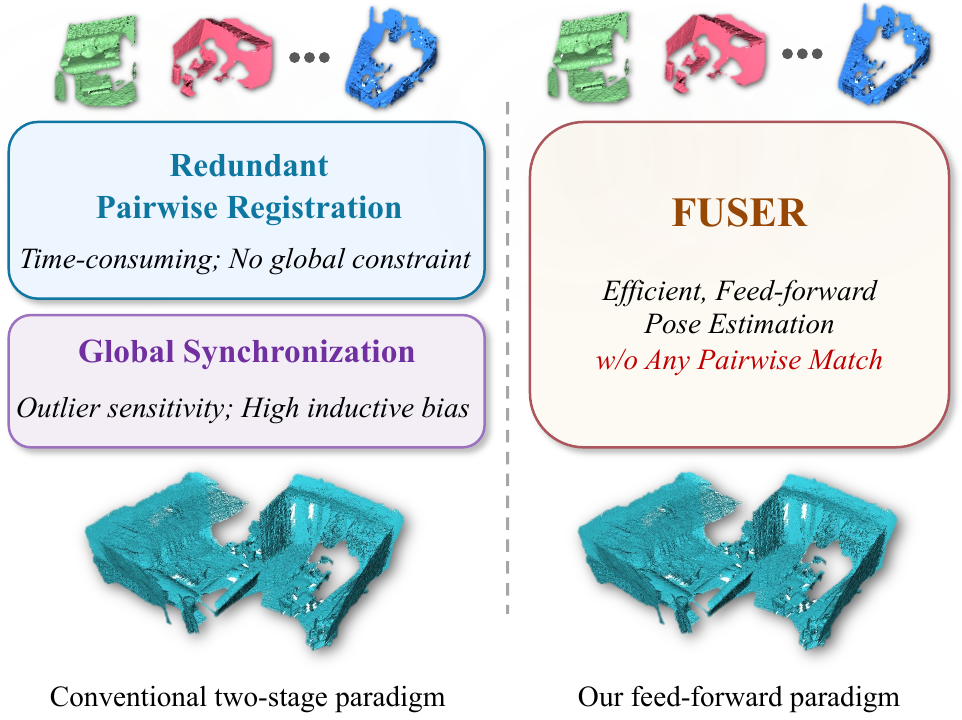}
	\vspace{-7mm}
	\caption{Comparison of paradigms. Conventional multiview registration relies on redundant pairwise estimation (time-consuming and no global constraint) and pose synchronization (outlier sensitivity and high inductive bias). By contrast, our FUSER directly predicts global poses through unified feed-forward reasoning across all scans without any pairwise matching, delivering outstanding accuracy and efficiency (minutes$\rightarrow$seconds).}
	\label{highlight}
	\vspace{-7mm}
\end{figure}

Despite its practical importance, the {\em multiview} registration problem has received far less attention compared to its \textit{pairwise} counterpart~\cite{zeng20173dmatch,huang2021predator,qin2022geometric,yu2023rotation}.
This is largely because most existing approaches adhere to a {\em pairwise-then-global} paradigm~\cite{arrigoni2016spectral,chatterjee2017robust,gojcic2020learning},
where pairwise registration serves as a sub-module to first estimate relative poses between pairs of scans (i.e.,  \textit{pose graph}), after which a pose-graph synchronization step is then applied to recover globally consistent absolute poses (i.e.,  \textit{transformation synchronization}). 
As the core component, pairwise registration has thus attracted most research effort, with the implicit assumption that improving pairwise accuracy will naturally bring better multiview consistency. 
However, optimizing scan pairs in isolation fails to exploit global multi-scan geometric constraints, often producing ambiguous or unstable relative poses whose errors further accumulate during synchronization. At the same time, registering many redundant pairs would impose substantial computational overhead.

To address these issues, recent studies move beyond exhaustive pairwise matching by prioritizing reliable scan pairs via overlap prediction~\cite{wang2023robust,cheng2024incremental,li2024matching} or triplet consistency~\cite{lee2022hara}.
Meanwhile, both analytical and learning-based synchronization frameworks have been developed to enhance global optimization.
Analytical methods employ robust averaging or hierarchical strategies to refine relative poses under geometric consistency~\cite{arrigoni2016spectral,wang2013exact,hartley2011l1,lee2022hara}, whereas learning-based approaches incorporate geometric priors and data-driven weighting to improve robustness~\cite{hu2024featsync,li2024matching,huang2019learning,gojcic2020learning,yew2021learning,jin2024multiway}.
Nevertheless, even the most advanced methods remain bound to the \textit{pairwise-then-global} paradigm and thus cannot fully overcome its inherent limitations.

In this paper, we propose FUSER, the first feed-forward multiview registration transformer that jointly encodes all scans in a compact latent space to directly predict global poses \textit{without any pairwise estimation}.
Unlike prior two-stage paradigm, FUSER enables holistic geometric reasoning across scans, avoids error propagation and redundant computation (caused by pairwise estimates), and removes handcrafted priors through a fully data-driven manner.

However, designing a feed-forward multiview registration transformer is highly non-trivial, as it must jointly reason across multiple scans while managing the prohibitive attention cost of all-scan interactions. To this end, we introduce absolute geometric encoding via a sparse 3D CNN~\cite{choy20194d,choy2019fully}, departing from relative encodings~\cite{thomas2019kpconv}, to encode each scan into low-resolution superpoint features that preserve absolute spatial cues crucial for absolute translation recovery. 
To ensure tractable multi-scan reasoning, we propose a Geometric Alternating Attention that alternates intra- and cross-scan message passing, capturing both local and global geometry. Notably, transferring 2D attention priors from an off-the-shelf VGGT-like foundation model (i.e., $\pi^3$~\cite{wang2025pi}) yields significant gains despite the modality gap. The fused features are decoded by a global pose predictor, trained with a relative pose loss as reference-free global supervision. 
Trained on large-scale datasets, FUSER achieves excellent accuracy while cutting runtime from minutes (in two-stage pipelines) to the second scale. 

Building on the established {FUSER} and its pose estimates, we further introduce {FUSER-DF}, an SE(3)$^N$ diffusion refinement framework that formulates multiview pose refinement as a denoising diffusion process in the joint SE(3)$^N$ space. We treat the {FUSER} outputs as pose priors, and design  prior-aware SE(3)$^N$ diffusion/reverse process to learn an SE(3)$^N$ denoiser that progressively refines the prior predictions toward fine-grained results. Remarkably, {FUSER} serves as a surrogate multiview registration model to construct the denoiser, and we also derive a prior-conditioned variational lower bound to supervise the denoising. 
Extensive experiments verify the outstanding accuracy and efficiency of our method.

To summarize, our main contributions are as follows:
\begin{itemize}
	\item We propose a novel \textit{feed-forward} paradigm for multi-view registration that directly predicts per-scan global poses \textit{without redundant pairwise matching}, enabling unified all-scan reasoning and reducing inference time from minutes to second scale.
	\vspace{1mm}
	\item We develop {\em FUSER}, an \textit{end-to-end multiview registration transformer} that combines absolute geometric encoding, 2D foundation attention-guided geometric alternating message passing, and direct pose regression, achieving 
	outstanding feed-forward accuracy and efficiency.
	\vspace{1mm}
	\item We introduce {FUSER-DF}, an {SE}(3)$^N$ diffusion refinement framework that further improves multiview pose estimates from {FUSER}. Here,  {FUSER} serves as a \textit{surrogate} multiview registration model to construct the {SE}(3)$^N$ denoiser, supervised by a prior-aware, multiview-specific variational lower bound. 
\end{itemize}

\section{Related Work}
\label{sec:relatedwork}
\noindent\textbf{Pairwise Registration.}
Pairwise registration~\cite{besl1992method,chetverikov2002trimmed,aiger20084,drost2010model,zhou2016fast} aims to estimate the relative transformation between two point clouds. 
Classical approaches use handcrafted descriptors to encode local geometry~\cite{johnson1999using,tombari2010unique,salti2014shot,rusu2008aligning,rusu2009fast}, such as USC~\cite{tombari2010unique} with a shape-context local frame, SHOT~\cite{salti2014shot} with normal-based histograms, and PFH/FPFH~\cite{rusu2008aligning,rusu2009fast} with pairwise geometric relations.
Recent learning-based descriptors adopt data-driven representations~\cite{zeng20173dmatch,wang2023zeroreg,bai2020d3feat,li2022lepard,li2020iterative,choy2020deep,chen2023sira,fu2021robust,ao2023buffer,jiang2023robust,jiang2025zero,jiang2025generative}.
Predator~\cite{huang2021predator} employs cross-attention to identify overlapping regions, RoITr~\cite{yu2023rotation} introduces rotation-invariant encoding, and GeoTransformer~\cite{qin2022geometric} injects explicit geometric embeddings for stronger discrimination. 
Other methods~\cite{shen2022reliable,jiang2021sampling,jiang2023center} are built on various paradigms such as reinforcement learning~\cite{liu2024MAM,liu2024OPT,liu2023CIA,jiang2022action,jiang2021action,zhao2021fast,jiang2021planning} and diffusion models~\cite{jiang2023se,wu2024diff,zhang2025diffpci}.
Beyond two-frame setting, we target the more challenging multiview alignment problem requiring globally consistent registration across multiple scans.

\noindent\textbf{Multiview Registration.}
Conventional multiview registration follows a two-stage pipeline: exhaustive pairwise registration followed by global pose synchronization.
Classical methods~\cite{arrigoni2016spectral,birdal2018bayesian,choi2015robust,govindu2004lie,huang2017translation} build fully connected pose graphs from all pairwise estimations, incurring quadratic complexity and degraded robustness under outliers~\cite{dong2018hierarchical,yew2021learning}.
To address these issues, traditional synchronization relies on non-linear least-squares frameworks such as IRLS~\cite{kummerle2011g,grisetti2011tutorial,clipp2010parallel}, often enhanced by robust rotation averaging~\cite{wang2013exact,hartley2011l1} or hierarchical initialization (e.g., HARA~\cite{lee2022hara}).
Recent learning-based approaches integrate geometric priors and data-driven robustness~\cite{gojcic2020learning,li2024matching,hu2024featsync,yew2021learning,jin2024multiway,huang2019learning,cheng2024incremental}.
For instance, LMVR~\cite{gojcic2020learning} and FeatSync~\cite{hu2024featsync} jointly optimize pairwise and global registration, SGHR~\cite{wang2023robust} predicts overlaps for adaptive reweighting, and MDGD~\cite{li2024matching} fuses descriptor and geometric cues for reliable graphs.
However, these methods still depend on extensive pairwise estimation and remain sensitive to noisy poses.
By contrast, our method eliminates the two-stage paradigm, directly predicting globally consistent poses in a single feed-forward process. 

\section{Approach}
\subsection{Background}\label{back}
\textbf{Problem Definition.} Given a collection of partially overlapping point clouds $\mathcal{S}=\{\mathbf{S}_i\in\mathbb{R}^{M_i\times 3}\mid 1\leq i\leq N\}$, each capturing a 3D scene fragment from a distinct camera pose, the goal of multiview 3D registration is to estimate the rigid global poses of all scans so that they can be consistently aligned within a common global coordinate system. 
Formally, for each scan $\mathbf{S}_i$, its rigid pose  is represented as $\mathbf{T}_i=(\mathbf{R}_i, \mathbf{t}_i)\in SE(3)$, 
where $\mathbf{R}_i\in SO(3)$ and $\mathbf{t}_i\in\mathbb{R}^3$ denote the rotation and translation of the $i$-th point cloud. 

\noindent\textbf{Existing Multiview Pipelines.}
In most existing multiview registration pipelines~\cite{choi2015robust,govindu2004lie,huang2017translation}, this problem is typically represented as a graph $\mathcal{G}=(\mathcal{S}, \mathcal{E})$, where each vertex corresponds to a scan and each edge $(i, j)\in\mathcal{E}$ encodes the relative pose $\mathbf{T}_{i\leftarrow j}=\mathbf{T}^{-1}_{i}\mathbf{T}_j$ between scans $\mathbf{S}_i$ and $\mathbf{S}_j$. 
The graph edges $\mathcal{E}$ are initialized from noisy relative transformations obtained by \textit{independent pairwise registration} algorithms~\cite{zeng20173dmatch,huang2021predator,qin2022geometric,yu2023rotation}.
A pose synchronization step is then applied to recover per-scan global poses. 

\noindent\textbf{Pairwise SE(3) Diffusion Registration.} 
The SE(3) diffusion registration model~\cite{jiang2023se} reformulates pairwise 3D registration as a denoising diffusion process composed of a forward SE(3) diffusion process and a reverse SE(3) denoising process. 
Given source and target point clouds $\mathcal{X}$ and $\mathcal{Y}$, the forward process gradually perturbs the optimal transformation $\mathbf{H}_0\in SE(3)$ into a noisy one, forming a Markov chain: $\mathbf{H}_0\rightarrow\mathbf{H}_1\cdots\rightarrow\mathbf{H}_T$, whose diffusion formula $\mathbf{H}_t\sim q(\mathbf{H}_t\mid \mathbf{H}_0)$ can be represented as: 
\begin{equation}
	\begin{split}
		\mathbf{H}_t=\operatorname{Exp}(\gamma\sqrt{1-\bar{\alpha}_t\boldsymbol{\varepsilon}})\mathcal{F}(\sqrt{\bar{\alpha}_t};\mathbf{H}_0, \mathbb{H}).
	\end{split}
\end{equation}
Here, Gaussian noise is sampled as $\boldsymbol{\varepsilon}\sim\mathcal{N}(\mathbf{0},\mathbf{I})$, and $\mathbb{H}$ denotes the identity transformation.  The diffusion coefficient is defined as $\bar{\alpha}_t=\prod_{s=0}^t\alpha_t=\prod_{s=0}^t(1-\beta_t)$, where $\beta_t$ follows a cosine schedule~\cite{nichol2021improved}. 
The {pose interpolation function} $\mathcal{F}(\sqrt{\bar{\alpha}_t};\mathbf{H}_0, \mathbb{H})=\operatorname{Exp}((1-\sqrt{\bar{\alpha}_t})\cdot\operatorname{Log}(\mathbb{H}\mathbf{H}_0^{-1}))\mathbf{H}_0$. 
The SE(3) reverse process then learns a pose denoiser to iteratively refine noisy poses through the chain  $\mathbf{H}_T\rightarrow\cdots\mathbf{H}_1\rightarrow\mathbf{H}_0$. 
Using a surrogate registration model $f_\theta$, the denoised pose can be estimated as $\mathbf{H}_{t-1} =$ 
\begin{equation}\label{diff0}
	\begin{split}
		\operatorname{Exp}(\lambda_0\operatorname{Log}(f_\theta(\mathcal{X}_t, \mathcal{Y})\mathbf{H}_t)+\lambda_1\operatorname{Log}(\mathbf{H}_t)+\sqrt{\tilde{\beta}_t}\boldsymbol{\varepsilon}),
	\end{split}
\end{equation}
where $\lambda_0=\frac{\sqrt{\bar{\alpha}_{t-1}}\beta_t}{1 - \bar{\alpha}_t}$ and $\lambda_1=\frac{\sqrt{\alpha_t}(1 - \bar{\alpha}_{t-1})}{1 - \bar{\alpha}_t}$, and  $\mathcal{X}_t$ denotes the source point cloud transformed by $\mathbf{H}_t$; Theoretically, the surrogate registration model can be modeled by different deep  pairwise registration models e.g.~\cite{wang2019deep,yew2020rpm,yew2022regtr}. 

\begin{figure}[t]
	\centering
	\includegraphics[width=\columnwidth]{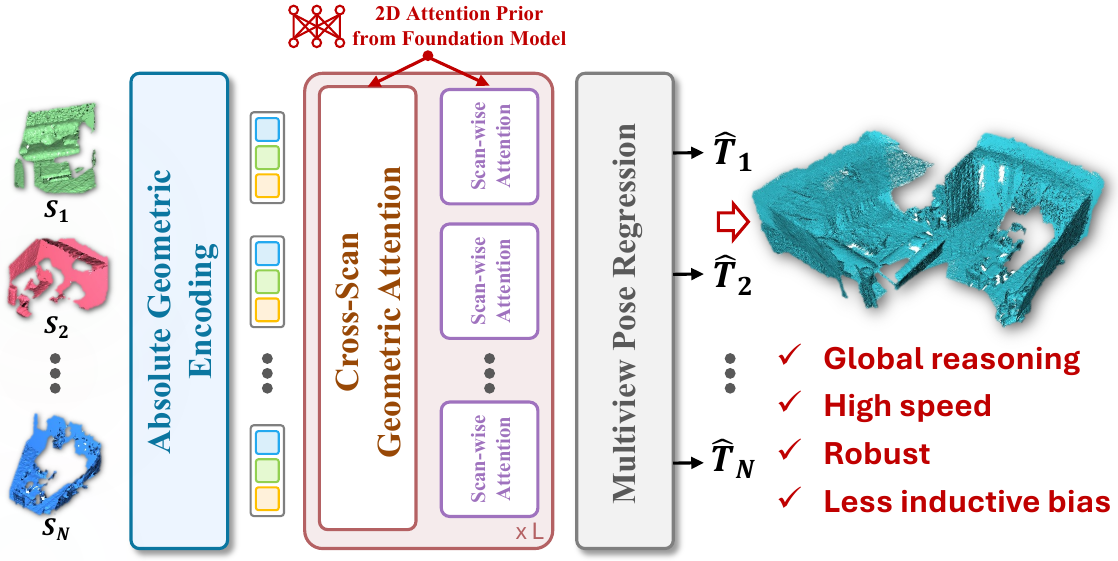}
	\vspace{-7mm}
	\caption{Architecture of FUSER. It encodes unordered scans into a compact latent space via \textit{Absolute Geometric Encoding}, then performs \textit{2D attention prior-enhanced Geometric Alternating Attention} for multiview reasoning and final pose regression.}
	\label{framework}
	\vspace{-5mm}
\end{figure}

\subsection{FUSER}
\subsubsection{Motivation}
Mainstream multiview registration systems still follow a \textit{pairwise-then-global} pipeline~\cite{arrigoni2016spectral,chatterjee2017robust,gojcic2020learning}: redundant pairwise alignments construct a pose graph, followed by pose synchronization on {SE}(3) to recover absolute poses.
While effective to some extent, this two-stage strategy has several fundamental limitations:
\textbf{(i) Missing global context:} Each pairwise alignment is performed independently, ignoring geometric constraints from other scans and leading to ambiguity in low-overlap or symmetric scenes that require joint reasoning;
\textbf{(ii) Outlier sensitivity.} Inaccurate pairwise estimates contaminate pose synchronization, causing global error propagation;
\textbf{(iii) Computational overhead:} Repeated pairwise registration is computationally intensive and markedly slow, due to costly feature extraction and outlier removal; 
\textbf{(iv) Strong inductive bias:} It necessitates numerous hand-crafted designs, e.g. graph sparsification, robust losses, and synchronization schedules, largely restricting model flexibility and hindering global optima. 

To address the aforementioned limitations, we propose {FUSER}, a novel, powerful and efficient feed-forward multiview 3D registration transformer that enables end-to-end global alignment without any time-consuming pairwise registration.  
FUSER takes all scans as input and performs geometric message passing across all views through stacked transformer layers in a compact latent space, directly regressing the absolute pose for each scan.  
By jointly reasoning over geometric cues from all scans, FUSER inherently mitigates the limitation \textbf{(i)} of lacking global context. Additionally, as the entire pipeline requires only a single, efficient feed-forward pass, without any redundant pairwise alignments, it effectively addresses the issues of outlier sensitivity and computational overhead in limitations \textbf{(ii)} and \textbf{(iii)}.  
Moreover, FUSER adopts a clean, fully data-driven architecture that alleviates the heavy reliance on handcrafted priors inherent in previous pipelines. It learns to infer accurate poses purely through transformer-based attention over raw point clouds, thereby alleviating the handcrafted inductive bias on the optimization space in limitation \textbf{(iv)}. 

As illustrated in Fig.~\ref{framework}, FUSER consists of an absolute geometric feature encoder, a 2D attention prior-enhanced alternating attention module, and a global pose predictor. The following subsections elaborate on each component.

\subsubsection{Absolute Geometric Encoding}\label{age}
Prevalent deep geometric descriptors~\cite{qin2022geometric,yu2023rotation} typically prefer translation-invariant descriptors, build on relative coordinate normalization \eg KPConv~\cite{thomas2019kpconv}, to stabilize pairwise matching.  
However, since {FUSER} directly regresses absolute poses, \textit{particularly translation}, such position-agnostic encodings become inadequate: removing global location cues obscures where each scan lies in the camera frame, making translation regression inherently ambiguous. 

To this end, we employ an \textit{absolute coordinate-aware} sparse 3D CNN built with MinkowskiEngine~\cite{choy20194d,choy2019fully} to extract hierarchical geometric features. 
Each scan $\mathbf{S}_i \in \mathcal{S}$ undergoes hierarchical voxelization and sparse convolutions, yielding compact, semantically enriched \textit{superpoints} $\mathbf{S}'_i \in \mathbb{R}^{M_i' \times 3}$ ($M_i' \ll M_i$) with features $\mathbf{F}_i \in \mathbb{R}^{M_i' \times d}$. 
To enable efficient all-scan attention in the transformer, we keep the latent representation highly compact by controlling the number of superpoints through a deeper, five-layer sparse convolutional hierarchy (kernel size 3, stride 2). Despite this low superpoint resolution, our experiments show that the resulting compact superpoint features still retain sufficient geometric information for accurate pose regression, leading to outstanding registration performance. 

% To enable efficient all-scan attention within the transformer (in the following subsection), we maintain highly compact latent features by limiting the superpoint count through deeper, total five sparse convolutional layers with (kernel size = 3, stride = 2).
% Despite the low superpoint resolution, empirical results show that our compact superpoint features can still encode effective geometric information for precise pose regression, achieving excellent registration accuracy.

\subsubsection{Geometric Alternating Attention} \label{aaatten}
With the downsampled superpoints $\mathcal{S}'=(\mathbf{S}'_1,...,\mathbf{S}'_{N})$ and their features $\mathcal{F}=(\mathbf{F}_1,...,\mathbf{F}_{N})$ from all scans, we then apply a transformer attention mechanism over them to enable comprehensive cross-scan geometric reasoning.

\noindent\textbf{Permutation-Equivariant Alternating Attention.} Inspired by VGGT’s alternating attention mechanism for cross-image reasoning~\cite{wang2025vggt}, we extend this paradigm to 3D matching field.  Our transformer contains $L=32$ alternating layers, including $16$ intra-scan and $16$ cross-scan blocks, applied sequentially to capture fine-grained local geometry and global contextual relationships across scans.

Notably, VGGT’s alternating attention uses learnable reference tokens to differentiate the reference view, but this design causes instability under varying scan orders.
To ensure permutation equivariance~\cite{wang2025pi}, we remove these tokens so that reordering scans does not alter their features:
\begin{equation}
	\operatorname{AA}(P_\pi(\mathcal{S}'), P_\pi(\mathcal{F})) = P_\pi(\operatorname{AA}(\mathcal{S}', \mathcal{F})),
\end{equation}
where $P_\pi(\cdot)$ denotes a permutation operator and $\operatorname{AA}(\cdot)$ the alternating attention. 
Additionally, unlike the 2D ROPE position embedding on 2D patch coordinates in VGGT, we replace it with sinusoidal positional encodings on superpoint coordinates to integrate absolute positional cues into the each attention layer. We note that we also devote much effort on relative position-aware 3D ROPE embedding, unfortunately, our experiments shows it may bring degraded performance, possibly caused by the relative position information would be misleading during cross-scan  attention (as each scan belongs to different coordinate system). 

\noindent\textbf{2D-to-3D Attention Prior Transfer.}
A key finding of our study is the cross-domain transfer of attention priors from 2D to 3D. Rather than training our alternating transformers from scratch, we investigate whether the powerful multiview reasoning learned by 2D reconstruction foundation models can benefit 3D geometric alignment.
To this end, we  initialize our alternating attention layers with pretrained weights from $\pi^3$, a VGGT variant trained on large-scale 2D image reconstruction tasks. 
Notably, we make no architectural modifications to accommodate this transfer due to their inherent structural compatibility (apart from differences in 2D/3D position embedding and 2D/3D feature encoding). 

This simple yet effective initialization empirically yields notable performance gains in our 3D multi-scan matching tasks as shown in Table~\ref{abla1}. We attribute these improvements to transferable 2D attention priors, such as view grouping, alignment consistency, and attention sparsity, which generalize surprisingly well to unstructured 3D point clouds. 
To the best of our knowledge, this demonstrates a promising step toward transferring pretrained 2D attention modules for 3D point cloud reasoning, suggesting a potential direction for future cross-modal foundation model research.

\subsubsection{Global Pose Prediction and Supervision}\label{loss}
\noindent\textbf{Global Pose Regression.}
With the superpoint tokens $\tilde{\mathcal{F}}=(\tilde{\mathbf{F}}_1,...,\tilde{\mathbf{F}}_{N})$ learned by  $L$ alternating attentions above, we directly regress a global pose $\hat{\mathbf{T}}_i\in SE(3)$ for each scan. 
Specifically, a series of self-attention blocks first refines the superpoint tokens of each scan, followed by global average pooling to obtain a compact scan-level global descriptor.
Two lightweight MLP heads then regress the translation $\hat{\mathbf{t}}_i \in \mathbb{R}^3$ and a $3{\times}3$ rotation proxy (a 9D representation), which is finally projected onto a valid rotation matrix $\hat{\mathbf{R}}_i \in SO(3)$ via SVD-based orthogonalization~\cite{wang2025pi}.

\noindent\textbf{Global Pose Supervision.}  
Directly supervising global poses in a fixed world frame is inherently ill-posed:
\textbf{(i)} world coordinate systems vary across sequences or datasets, leading to instability when regressing absolute ground-truth poses; and
\textbf{(ii)} defining the world frame by arbitrarily selecting a reference scan breaks permutation equivariance with respect to scan order.
To address these issues, we adopt a reference-free strategy that indirectly supervises the \emph{pairwise relative poses}. 
Specifically, for any $(i\!\neq\! j)$, we compute the relative transformation $\hat{\mathbf{T}}_{i\leftarrow j}=\hat{\mathbf{T}}_i^{-1}\,\hat{\mathbf{T}}_j$ and compare it to the ground-truth $\mathbf{T}_{i\leftarrow j}$ using a \textit{geodesic rotation loss} and a \textit{robust translation loss} with Huber function $\ell_\beta(\cdot)$: 
\begin{equation}\label{pointloss}
	\begin{split}
		\mathcal{L}_{\mathbf{r}}(i, j) &= \operatorname{arccos}\left(\frac{\operatorname{Tr}(\mathbf{R}_{i\leftarrow j}^{\top}\hat{\mathbf{R}}_{i\leftarrow j})-1}{2}\right), \\
		\mathcal{L}_{\mathbf{t}}(i, j) &= \ell_\beta\left(\hat{\mathbf{t}}_{i\leftarrow j}-\mathbf{t}_{i\leftarrow j}\right). 
	\end{split}
\end{equation}
We further impose a \textit{point-wise loss} to ensure geometric consistency between the transformed point clouds under the predicted and ground-truth poses $\mathcal{L}_{\mathbf{p}}(i,j) =$
\begin{equation}\small
	\begin{split}
		\frac{1}{N_j}\sum_{l=1}^{N_j} \|(\hat{\mathbf{R}}_{i\leftarrow j}\mathbf{p}_l+\hat{\mathbf{t}}_{i\leftarrow j})-({\mathbf{R}}_{i\leftarrow j}\mathbf{p}_l+{\mathbf{t}}_{i\leftarrow j})\|_1.
	\end{split}
\end{equation}
Finally, our overall loss is summarized as $\mathcal{L} =$
\begin{equation}\label{loss2}
	\begin{split}
		\frac{1}{N(N -1)}\sum_{i\neq j}\mathcal{L}_\mathbf{r}(i,j) + \gamma_t\mathcal{L}_\mathbf{t}(i,j) + \gamma_p\mathcal{L}_\mathbf{p}(i,j),
	\end{split}
\end{equation}
where $\gamma_t$ and $\gamma_p$ balance the respective loss terms.

\subsection{FUSER-DF: SE(3)$^N$ Diffusion Refinement} 
Building on our established feed-forward multiview registration transformer (FUSER) and its predicted global poses $\hat{\mathbf{T}}_{1:N}=(\hat{\mathbf{T}}_{1},...,\hat{\mathbf{T}}_{N})$, we further seek higher alignment precision by introducing a novel {SE}(3)$^N$ Diffusion Refinement model, termed {FUSER-DF}, that formulates the multiview pose correction as a denoising diffusion process over the joint {SE}(3)$^N$ manifold. 

\subsubsection{FUSER-DF vs. SE(3) Diffusion Registration} 
Compared with the prior SE(3) diffusion model~\cite{jiang2023se}, our {FUSER-DF} brings three key innovations:

\noindent\textbf{(i)} {From pairwise to multiview:} Prior work diffuses a single relative motion on SE(3), whereas {FUSER-DF} diffuses the {entire pose set} on the joint space SE(3)$^{N}$, preserving cross-scan dependencies throughout refinement;  

\noindent\textbf{(ii)} {From estimation to refinement:} Unlike denoising from an uninformed identity prior $\mathbb{H}$ for pose estimation, our pose refinement initializes the reverse chain at {FUSER}’s pose estimate and only need perform {small-step} correction;  

\noindent\textbf{(iii)} {Pairwise to multi-view surrogate:} Prior work relies on a {pairwise} surrogate registration model $f_\theta$ and thus cannot model multiview alignment; Thanks to the multiview pose estimation of {FUSER}, we can employ it as our multiview surrogate registration model for SE(3)$^{N}$ refinement. 

\subsubsection{Prior-aware {SE}(3)$^{N}$ Diffusion Process}
Unlike SE(3) diffusion~\cite{jiang2023se} that perturbs optimal transformation toward the identity one, our prior-aware SE(3)$^N$ diffusion instead diffuses from the optimal poses toward the prior $\hat{\mathbf{T}}_{1:N}=(\hat{\mathbf{T}}_{1},...,\hat{\mathbf{T}}_{N})$ predicted by {FUSER}. This process forms a Markov chain $\mathbf{T}_{1:N}^{0}\rightarrow\mathbf{T}_{1:N}^{1}\!\cdots\!\mathbf{T}_{1:N}^{T}$, where the diffusion formula $\mathbf{T}_{1:N}^{t}=(\mathbf{T}_{1}^{t},...,\mathbf{T}_{N}^{t})\sim q(\mathbf{T}_{1:N}^{t}\mid \mathbf{T}_{1:N}^{0}, \hat{\mathbf{T}}_{1:N})$ can be expressed as:
\begin{equation}\label{diff2}
	\begin{split}
		\mathbf{T}_i^{t}={\operatorname{Exp}(\gamma\sqrt{1-\bar{\alpha}_t\boldsymbol{\varepsilon}})}{\mathcal{F}(\sqrt{\bar{\alpha}_t};\mathbf{T}_i^{0}, \hat{\mathbf{T}}_i)},
	\end{split}
\end{equation}
where $i\in\{1,...,N\}$ denotes the scan index; $\mathcal{F}$ represents the pose interpolation function between the optimal pose and the prior pose estimate, with $\sqrt{\bar{\alpha}_t}$ acting as the interpolation weight: $\mathcal{F}(\sqrt{\bar{\alpha}_t};\mathbf{T}_i^{0}, \hat{\mathbf{T}}_i)=$
\begin{equation}
	\begin{split}
		\operatorname{Exp}((1-\sqrt{\bar{\alpha}_t})\cdot\operatorname{Log}(\hat{\mathbf{T}}_i{(\mathbf{T}_i^{0})}^{-1}))\mathbf{T}_i^{0}.
	\end{split}
\end{equation}
The perturbation $\operatorname{Exp}(\gamma\sqrt{1-\bar{\alpha}_t\boldsymbol{\varepsilon}})$ introduces stochastic noise to the interpolated pose, enriching the diversity of diffusion trajectories. 
This diversity effectively augments the training of the denoiser network, enhancing its generalization capability across diverse pose distributions.

\subsubsection{Prior-aware {SE}(3)$^{N}$ Reverse Process}
\begin{figure}[t]
	\centering
	\includegraphics[width=\columnwidth]{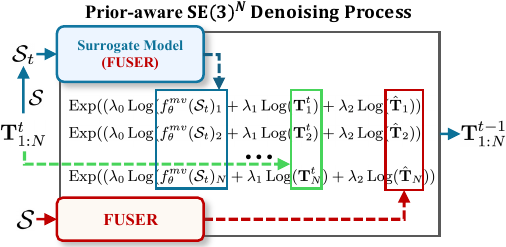}
	\vspace{-5mm}
	\caption{Pipeline of prior-aware SE(3)$^N$ denoising process. It integrates the prior pose estimates $(\hat{\mathbf{T}}_1,...,\hat{\mathbf{T}}_N)$ of FUSER into the denoising process, where FUSER, as the surrogate registration model, estimates the residual poses $(\hat{\mathbf{T}}_1^{t\rightarrow 0}, ..., \hat{\mathbf{T}}_N^{t\rightarrow 0})=\operatorname{\textcolor[RGB]{166,35,35}{\textbf{FUSER}}}(\mathcal{S}_t)$ to support progressive denoising over SE(3)$^N$ space.}
	\label{denoise}
	\vspace{-5mm}
\end{figure}

As demonstrated in Sec.~\ref{back}, the conventional SE(3) reverse process learns an SE(3) denoiser that refines poses from the identity transformation to the optimal one. By contrast, our prior-aware SE(3)$^N$ reverse process learns an SE(3)$^N$ denoiser $p_\theta$ that refines poses from the prior multiview poses predicted by {FUSER} toward the optimal ones (see Fig.~\ref{denoise}). 
To train this denoiser, we first derive a prior-aware variational lower bound of  log-likelihood over the training data:
\begin{equation}\small \label{vlb}
	\begin{split}
		\setlength{\abovedisplayskip}{2pt}
		\setlength{\belowdisplayskip}{2pt}
		& \mathbb{E}_{\mathcal{S}, \mathbf{T}^0_{1:N}\sim p_{data}}\left[\ln p_\theta(\mathbf{T}^0_{1:N}\mid\mathcal{S}, \hat{\mathbf{T}}_{1:N}) \right] \\
		\geq & \mathbb{E}\left[\ln\frac{p_\theta(\mathbf{T}^{0:T}_{1:N}\mid \mathcal{S}, \hat{\mathbf{T}}_{1:N})}{q(\mathbf{T}_{1:N}^{1:T}\mid \mathbf{T}_{1:N}^0, \hat{\mathbf{T}}_{1:N})}\right]=\underbrace{\mathbb{E}\left[\ln{p_\theta(\mathbf{T}_{1:N}^0\mid\mathcal{S}, \hat{\mathbf{T}}_{1:N})}\right]}_{\text{Residual term}} \\
		& - \underbrace{\mathbb{E}\left[\operatorname{{KL}}(q(\mathbf{T}_{1:N}^T\mid \mathbf{T}_{1:N}^0, \hat{\mathbf{T}}_{1:N}) || p(\mathbf{T}_{1:N}^T\mid\hat{\mathbf{T}}_{1:N}))\right]}_{\text{Prior matching term}} \\
		- & \mathbb{E}\Big[\sum^{T}_{t=2}\underbrace{\operatorname{{KL}}(q(\mathbf{T}_{1:N}^{t-1}\mid \mathbf{T}_{1:N}^t, \mathbf{T}_{1:N}^0, \hat{\mathbf{T}}_{1:N}) || {p_\theta(\mathbf{T}_{1:N}^{t-1}\mid\mathcal{S}_t, \hat{\mathbf{T}}_{1:N})})}_{\text{Prior-aware Denoising matching term}} \Big],
	\end{split}
\end{equation}
where $\mathcal{S}_t$ represents the transformed scans under the noisy poses $\mathbf{T}^t_{1:N}$, that is  $\mathcal{S}_t=\{\mathbf{S}_i^t\mid\mathbf{S}_i^t=\mathbf{R}_i^t\mathbf{S}_i+\mathbf{t}_i^t, i=1..N\}$.
Please refer to Appendix B for detailed derivation. 

\noindent\textbf{Prior-aware Denoising Matching Term.} This term serves as the core loss for training our SE(3)$^N$ denoiser. Here, $q(\mathbf{T}_{1:N}^{t-1}\mid \mathbf{T}_{1:N}^t, \mathbf{T}_{1:N}^0, \hat{\mathbf{T}}_{1:N})$ denotes the posterior distribution of noisy poses $\mathbf{T}_{1:N}^{t-1}$, while ${p_\theta(\mathbf{T}_{1:N}^{t-1}\mid\mathcal{S}_t, \hat{\mathbf{T}}_{1:N})}$ represents its learnable prior distribution predicted by the denoiser. Benefiting from the pose reference from both the optimal poses $\mathbf{T}_{1:N}^0$ and the prior poses $\hat{\mathbf{T}}_{1:N}$, the posterior can naturally provide a supervision signal for guiding the training of the ground truth-agnostic prior distribution  by minimizing their Kullback-Leibler (KL) divergence. Formally, inspired by~\cite{han2022card}, our prior-aware posterior pose, ${}^{post}\mathbf{T}_{1:N}^{t-1}\sim q(\mathbf{T}_{1:N}^{t-1}\mid \mathbf{T}_{1:N}^t, \mathbf{T}_{1:N}^0, \hat{\mathbf{T}}_{1:N})$, can be expressed as: ${}^{post}\mathbf{T}_{i}^{t-1}=$
\begin{equation}\small\label{vae11prior2}
	\begin{split}
		\operatorname{Exp}(\lambda_0 \operatorname{Log}{(\mathbf{T}_{i}^{0})}+\lambda_1\operatorname{Log}{(\mathbf{T}_{i}^{t})} +\lambda_2\operatorname{Log}(\hat{\mathbf{T}}_{i})+\tilde{\boldsymbol{\varepsilon}}_i). 
	\end{split}
\end{equation}
Here, $\tilde{\boldsymbol{\varepsilon}}_i=\sqrt{\tilde{\beta}_t} \boldsymbol{\varepsilon}_i$ and  the scan index $i\in\{1,...,N\}$; 
The denoising coefficients $\lambda_0$ and $\lambda_1$ are consistent with those in Eq.~\ref{diff0}, and $\lambda_2= 1 + \frac{(\sqrt{\bar{\alpha}_t} - 1)(\sqrt{{\alpha}_t} + \sqrt{\bar{\alpha}_{t-1}})}{1 - \bar{\alpha}_t} $. 
The denoising coefficients $\lambda_0$, $\lambda_1$, and $\lambda_2$ essentially control the relative contributions of the optimal pose $\mathbf{T}_{i}^{0}$, the current noisy pose $\mathbf{T}_{i}^{t}$, and the prior pose estimate $\hat{\mathbf{T}}_{i}$ at each timestep.
Intuitively, at the beginning of the reverse process, $\lambda_2$ dominates, making the posterior pose strongly depend on the prior estimate. As the denoising progresses, $\lambda_2$ gradually decays, shifting the posterior toward the conventional SE(3) reverse formulation governed by $\lambda_0$ and $\lambda_1$. Consequently, following the formulation in~\cite{jiang2023se}, the expected value of posterior pose (Eq.~\ref{vae11prior2}) can be rewritten as $\mu^{post}=$ 
\begin{equation}\small\label{vae11prior2}
	\begin{split}
		\operatorname{Exp}(\lambda_0 \operatorname{Log}{(\mathbf{T}_{i}^{t\rightarrow 0}\mathbf{T}_{i}^{t})}+\lambda_1\operatorname{Log}{(\mathbf{T}_{i}^{t})} +
		\lambda_2\operatorname{Log}(\hat{\mathbf{T}}_{i})),
	\end{split}
\end{equation} 
where $\mathbf{T}_{i}^{t\rightarrow 0}=\mathbf{T}_{i}^{0}(\mathbf{T}_{i}^{t})^{-1}$ represents the residual transformation from the noisy pose to the optimal one.
Analogously, the expected value of the parameterized prior distribution can be reformulated through a multiview surrogate registration model $f_\theta^{mv}$ as
$\mu_\theta(\mathcal{S}, \mathbf{T}_{1:N}^{t}, \hat{\mathbf{T}}_{1:N})_i\triangleq$
\begin{equation}\label{meanprior}\small
	\begin{split}
		\text{Exp}((\lambda_0\text{Log}(f_\theta^{mv}(\mathcal{S}_t)_i\mathbf{T}_{i}^{t} +\lambda_1\text{Log}(\mathbf{T}_i^t) + \lambda_2\text{Log}(\hat{\mathbf{T}}_i)). 
	\end{split}
\end{equation}
Benefiting from our established end-to-end feed-forward multiview registration model, {FUSER}, we employ it to instantiate the multiview surrogate function as:
\begin{equation}\label{surro}\small
	\begin{split}
		(\hat{\mathbf{T}}_1^{t\rightarrow 0}, \hat{\mathbf{T}}_2^{t\rightarrow 0}, ..., \hat{\mathbf{T}}_N^{t\rightarrow 0})=f_\theta^{mv}(\mathcal{S}_t)\triangleq\operatorname{\textcolor[RGB]{166,35,35}{\textbf{FUSER}}}(\mathcal{S}_t).
	\end{split}
\end{equation}
Minimizing the KL divergence between the prior and the posterior is thus equivalent to reducing their expected-value discrepancy (Eq.~\ref{vae11prior2} and Eq.~\ref{meanprior}). Consequently, this objective can be reformulated as optimizing the surrogate model $f_\theta^{mv}(\mathcal{S}_t)$ to predict the residual pose $\mathbf{T}_{i}^{t\rightarrow 0}$, a formulation naturally aligned with the standard optimization target of multiview registration. Thus, the losses introduced in Sec.~\ref{loss} can be directly applied for surrogate training.

\noindent\textbf{Residual and Prior Matching Terms}. The prior matching term actually is a constant term and doesn't contain any optimizable parameters, and we can thus ignore it. Instead, the objective of residual matching term is essentially consistent with the prior-aware denoising matching term at $t=1$. 

\begin{table*}[h]
	\centering
	\caption{Multiview registration performance on the  \textbf{ScanNet (30 scans)}~\cite{dai2017scannet}. }
	\vspace{-3mm}
	\resizebox{1\textwidth}{!}{
		\begin{tabular}{c l ccccccccccccccc}
			\toprule[1.8pt]
			\multirow{3}{*}{\textbf{Pose Graph}} & \multirow{3}{*}{\textbf{Method}} & \multirow{3}{*}{\textbf{\#Pair}}  & \multicolumn{6}{c}{\textbf{Rotation Error}} & \multicolumn{6}{c}{\textbf{Translation Error} (\textit{m})} \\
			\cmidrule(lr){4-9} \cmidrule(lr){10-15} 
			&  & & 3$^\circ$ & 5$^\circ$ & 10$^\circ$ & 30$^\circ$ & 45$^\circ$ & Mean/Med & 0.05 & 0.1 & 0.25 & 0.5 & 0.75 & Mean/Med  \\
			\midrule[1.2pt]
			\multirow{8}{*}{\textbf{Full}} & LMVR~\cite{gojcic2020learning}  & 13920 & 48.3 & 53.6 & 58.9 & 63.2 & 64.0 & 48.1$^\circ$/33.7$^\circ$ & 34.5 & 49.1 & 58.5 & 61.6 & 63.9 & 0.83/0.55 \\ 
			%                & {MST}~\cite{huber2003fully} & 13920 & 28.2 & 42.6 & 54.4 & 70.4 & 73.9 & {34.3$^\circ$/23.6$^\circ$} & 27.4 & 45.8 & 63.2 & 66.0 & 69.4 & {0.82/0.65} \\
			& EIGSE3~\cite{arrigoni2016spectral}  &  13920 & 19.7 & 24.4 & 32.3 & 49.3 & 56.9 & 53.6$^\circ$/48.0$^\circ$ & 11.2 & 19.7 & 30.5 & 45.7 & 56.7 & 1.03/0.94 \\
			& L1-IRLS~\cite{chatterjee2017robust}  & 13920 & 38.1 & 44.2 & 48.8 & 55.7 & 56.5 & 53.9$^\circ$/47.1$^\circ$ & 18.5 & 30.4 & 40.7 & 47.8 & 54.4 & 1.14/1.07 \\
			& RotAvg~\cite{chatterjee2017robust} &13920 & 44.1 & 49.8 & 52.8 & 56.5 & 57.3 & 53.1$^\circ$/44.0$^\circ$ & 28.2 & 40.8 & 48.6 & 51.9 & 56.1 & 1.13/1.05 \\ 
			& LITS~\cite{yew2021learning} &  13920 & 52.8 & 67.1 & 74.9 & 77.9 & 79.5 & 26.8$^\circ$/27.9$^\circ$ & 29.4 & 51.1 & 68.9 & 75.0 & 77.0 & 0.68/0.66 \\
			& HARA~\cite{lee2022hara} & 13920 & 54.9 & 64.3 & 71.3 & 74.1 & 74.2 & 32.1$^\circ$/29.2$^\circ$ & 35.8 & 54.4 & 66.3 & 69.7 & 72.9 & 0.87/0.75 \\
			& SGHR~\cite{wang2023robust} & 13920 & {57.2} & {68.5} & {75.1} & {78.1} & {78.8} & {26.4$^\circ$/19.5$^\circ$} & {39.4} & {61.5} & {72.0} & {75.2} & {77.6} & {0.70/0.59} \\
			& MDGD~\cite{li2024matching} & 13920 & {54.7} & {71.4} & {83.4} & {88.2} & {88.6} & {17.6$^\circ$/19.1$^\circ$} & {38.7} & {62.8} & {77.8} & {82.6} & {85.2} & {0.42/0.35} \\ 
			\midrule[1pt]
			\multirow{6}{*}{\textbf{Pruned}~\cite{gojcic2020learning}} & EIGSE3~\cite{arrigoni2016spectral}  & 13920 & 40.8 & 46.3 & 51.9 & 61.2 & 65.7 & 40.6$^\circ$/37.1$^\circ$ & 23.9 & 38.5 & 51.0 & 59.3 & 66.1 & 0.88/0.84 \\ 
			& L1-IRLS~\cite{chatterjee2017robust}  & 13920 & 46.3 & 54.2 & 61.6 & 64.3 & 66.8 & 41.8$^\circ$/34.0$^\circ$ & 24.1 & 38.5 & 48.3 & 55.6 & 60.9 & 1.05/1.01 \\
			& RotAvg~\cite{chatterjee2017robust} & 13920 & 50.2 & 60.1 & 65.3 & 66.8 & 68.8 & 38.5$^\circ$/31.6$^\circ$ & 31.8 & 49.0 & 58.8 & 63.3 & 65.6 & 0.96/0.83 \\ 
			& LITS~\cite{yew2021learning} & 13920 & 54.3 & 69.4 & 75.6 & 78.5 & 80.3 & 24.9$^\circ$/19.9$^\circ$ & 31.4 & 54.4 & 72.3 & 76.7 & 79.6 & 0.65/0.56 \\
			& HARA~\cite{lee2022hara} & 13920 & 55.7 & 63.7 & 69.0 & 70.8 & 72.1 & 34.7$^\circ$/31.3$^\circ$ & 35.2 & 53.6 & 65.4 & 68.6 & 71.7 & 0.86/0.71 \\
			& SGHR~\cite{wang2023robust} & 13920 & {59.4} & {71.9} & {80.0} & {82.1} & {82.6} & {21.7$^\circ$/19.1$^\circ$} & {39.9} & {63.0} & {74.3} & {77.6} & {80.2} & {0.64/0.47} \\ 
			\midrule[1pt]
			\multirow{3}{*}{\textbf{Sparse}} & SGHR~\cite{wang2023robust} & {6004} & {59.1} & {73.1} & {80.8} & {82.5} & {83.0} & {21.7$^\circ$/19.0$^\circ$} 
			& {39.9} & {64.1} & {76.7} & {79.0} & {81.9} & {0.56/0.49}  \\
			& IncreMVR~\cite{cheng2024incremental} & {6004} & 58.6 & 73.4 & 79.7 & 83.9 & 85.6 & 19.8$^\circ$/15.6$^\circ$ & 39.6 & 63.9 & 76.3 & 81.9 & 85.4 & 0.55/0.37  \\
			& MDGD~\cite{li2024matching} & {6004} & 56.1 & 71.8 & {83.5} & 88.5 & 88.8 & 17.4$^\circ$/19.0$^\circ$ & 38.2 & 61.2 & 77.5 & 82.7 & 84.9 & 0.37/{0.31} \\ 
			\midrule[1pt] 
			% \multirow{2}{*}{/} & \textcolor[RGB]{166,35,35}{\textbf{FUSER}} (\textbf{Ours}) & \textbf{0} & {65.3} &  {85.9} &  \textbf{94.1} &  \textbf{95.4} &  \textbf{95.9} &  \textbf{7.0}$^\circ$/\textbf{6.6}$^\circ$ & 32.3 & {65.1} & {91.2} & \textbf{95.1} & \textbf{96.2} &  \textbf{0.15}/{0.14}\\ 
			% & \textcolor[RGB]{166,35,35}{\textbf{FUSER-DF}} (\textbf{Ours})& \textbf{0} & \textbf{71.8} & \textbf{88.8} & \textbf{94.1} & {94.8} & {95.1} &  {7.5}$^\circ$/{7.3}$^\circ$ & \textbf{40.4} & \textbf{71.7} & \textbf{92.2} & {94.5} & {95.3} &  {0.16}/\textbf{0.13} \\ 
            \multirow{2}{*}{/} & \textcolor[RGB]{166,35,35}{\textbf{FUSER}} (\textbf{Ours}) & \textbf{0} & {69.4} &  {86.9} &  \textbf{94.7} &  \textbf{95.8} &  \textbf{96.1} &  \textbf{6.7}$^\circ$/{2.1}$^\circ$ & 36.2 & {71.8} & {92.5} & \textbf{95.4} & \textbf{96.1} &  \textbf{0.15}/{0.07}\\ 
			& \textcolor[RGB]{166,35,35}{\textbf{FUSER-DF}} (\textbf{Ours})& \textbf{0} & \textbf{72.0} & \textbf{89.7} & {94.5} & {95.2} & {95.5} &  {7.1}$^\circ$/\textbf{2.0}$^\circ$ & \textbf{43.2} & \textbf{75.5} & \textbf{92.8} & {94.7} & {95.6} &  \textbf{0.15}/\textbf{0.06} \\ 
			\bottomrule[1.8pt]
	\end{tabular}}
	\label{scannetcompa}
	\vspace{-4mm}
\end{table*}

\begin{table}[t]
	\centering
	\caption{Comparisons on \textbf{3DMatch (60 scans)}~\cite{zeng20173dmatch}. }
	\vspace{-3mm}
	\resizebox{1\columnwidth}{!}{
		\begin{tabular}{lcccc}
			\toprule[1.8pt]
			{\textbf{Method}} & \textbf{\#Pair} & \textbf{RR} (\%) $\uparrow$  & \textbf{RE} ($^\circ$) $\downarrow$  &  \textbf{TE} (m) $\downarrow$   \\
			\midrule[1.2pt]
			Full+FCGF~\cite{choy2019fully}   & 13615 & 46.2 & 33.8 & 0.89   \\ 
			Full+Predator~\cite{huang2021predator}  & 13615 & 49.7 & 26.8 & 0.67   \\
			Full+YOHO~\cite{wang2022you}   &  13615 & 59.2 & 35.8 &  0.77 \\
			Full+GeoTrans~\cite{qin2022geometric}   &  13615 & 61.5 & 38.0 & 0.61 \\
			Full+PARENet~\cite{yao2024pare}   &  13615 & 61.9 & 31.2 & 0.68 \\
			\midrule[1pt]
			SGHR+FCGF~\cite{choy2019fully}   & 3034 &  39.7 & 38.4 & 0.66 \\ 
			SGHR+Predator~\cite{huang2021predator}  & 3034 & 44.2 & 28.3 &  0.55   \\
			SGHR+YOHO~\cite{wang2022you}   &  3034 & 53.2 & 34.6 & 0.57 \\
			SGHR+GeoTrans~\cite{qin2022geometric}   &  3034 & 55.2 & 29.8 & 0.69 \\
			SGHR+PARENet~\cite{yao2024pare}   &  3034 & 55.3 & 37.0 & 0.60  \\ 
			\midrule[1pt]
			% \textcolor[RGB]{166,35,35}{\textbf{FUSER}} (\textbf{Ours}) & \textbf{0} & 88.7 & 3.2 &  0.16  \\
			% \textcolor[RGB]{166,35,35}{\textbf{FUSER-DF}} (\textbf{Ours})& \textbf{0} & \textbf{90.7} & \textbf{2.9} & \textbf{0.15}  \\
            			\textcolor[RGB]{166,35,35}{\textbf{FUSER}} (\textbf{Ours}) & \textbf{0} & 90.3 & 3.2 &  \textbf{0.14}  \\
			\textcolor[RGB]{166,35,35}{\textbf{FUSER-DF}} (\textbf{Ours})& \textbf{0} & \textbf{92.0} & \textbf{3.1} & \textbf{0.14}  \\
			\bottomrule[1.8pt]
	\end{tabular}}
	\label{3dmatchcompa}
	\vspace{-1mm}
\end{table}

\begin{table}[t]
	\centering
	\caption{Comparisons on \textbf{ArkitScenes (200 scans)}~\cite{baruch2021arkitscenes}. }
	\vspace{-3mm}
	\resizebox{1\columnwidth}{!}{
		\begin{tabular}{lcccc}
			\toprule[1.8pt]
			{\textbf{Method}} & \textbf{\#Pair} & \textbf{RR} (\%) $\uparrow$  & \textbf{RE} ($^\circ$) $\downarrow$  &  \textbf{TE} (m) $\downarrow$   \\
			\midrule[1.2pt]
			% Full+FCGF~\cite{choy2019fully}   & 207527   \\ 
			% Full+Predator~\cite{huang2021predator}  & 207527   \\
			% Full+YOHO~\cite{wang2021you}   & 207527   \\
			Full+GeoTrans~\cite{qin2022geometric}   & 207527 & 24.7 & 92.5 & 1.36 \\
			Full+PARENet~\cite{yao2024pare}   & 207527 & 14.3 & 109.2 & 1.44  \\
			\midrule[1pt]
			% SGHR+FCGF~\cite{choy2019fully}   & 47924 &   \\ 
			% SGHR+Predator~\cite{huang2021predator}  & 47924 &   \\
			% SGHR+YOHO~\cite{wang2021you}   &  47924 & \\
			SGHR+GeoTrans~\cite{qin2022geometric}   &  47924 & 26.7 & 91.2 & 1.48 \\
			SGHR+PARENet~\cite{yao2024pare}   &  47924 &  23.7 & 95.3 & 1.24 \\ 
			\midrule[1pt]
			% \textcolor[RGB]{166,35,35}{\textbf{FUSER}} (\textbf{Ours}) & \textbf{0} & 84.0 & \textbf{6.7} &  0.20  \\
			% \textcolor[RGB]{166,35,35}{\textbf{FUSER-DF}} (\textbf{Ours})& \textbf{0} & \textbf{89.1} & {7.2} & \textbf{0.18}  \\
            \textcolor[RGB]{166,35,35}{\textbf{FUSER}} (\textbf{Ours}) & \textbf{0} & 92.1 & \textbf{5.4} &  0.13  \\
			\textcolor[RGB]{166,35,35}{\textbf{FUSER-DF}} (\textbf{Ours})& \textbf{0} & \textbf{95.0} & {5.6} & \textbf{0.12}  \\
			\bottomrule[1.8pt]
	\end{tabular}}
	\label{ark}
	\vspace{-5mm}
\end{table}

\begin{table*}[h]
	\centering
	\caption{Ablation Studies on \textbf{ScanNet}~\cite{dai2017scannet} (\textit{ScaN}: ScanNet~\cite{dai2017scannet}, \textit{ArkitS}: ArkitScenes~\cite{baruch2021arkitscenes}, \textit{ScaNP}: ScanNet++~\cite{yeshwanth2023scannet++}, \textit{3DM}:  3DMatch~\cite{zeng20173dmatch}).}
	\vspace{-3mm}
	\resizebox{1\textwidth}{!}{
		\begin{tabular}{clcccccccccccccc}
			\toprule[1.8pt]
			\multirow{3}{*}{\textbf{Pose Graph}} & \multirow{3}{*}{\textbf{Method}}& \multicolumn{6}{c}{\textbf{Rotation Error}} & \multicolumn{6}{c}{\textbf{Translation Error} (\textit{m})} \\
			\cmidrule(lr){3-8} \cmidrule(lr){9-14} 
			&  & 3$^\circ$ & 5$^\circ$ & 10$^\circ$ & 30$^\circ$ & 45$^\circ$ & Mean/Med & 0.05 & 0.1 & 0.25 & 0.5 & 0.75 & Mean/Med  \\
			\midrule[1.2pt]
			\multirow{5}{*}{\textbf{FUSER}} & (ScaN) \textit{w/o} 2D Attention Prior & 12.9 & 26.8 & 46.9 & 67.3 & 73.4 & 34.8$^\circ$/11.3$^\circ$ &  5.4 & 16.8 & 37.8 & 56.8 & 68.4 &  0.74/0.40  \\
			& (ScaN) & 36.6 & 57.3 & 73.9 & 83.5 & 85.1 & 22.6$^\circ$/4.1$^\circ$ & 15.4 & 36.5 & 65.0 & 80.7 & 84.3 &  0.45/0.16 \\
			& (ScaN + ArkitS) &  35.8 & 63.0 & 87.0 & 93.0 & 94.0 & 11.4$^\circ$/3.9$^\circ$ & 16.3 & 44.8 & 80.5 & 90.9 & 93.2 &  0.26/0.11 \\
			& (ScaN + ArkitS + ScaNP) & 56.9 & 79.8 & 91.5 & 93.1 & 93.6 &  9.4$^\circ$/\textbf{2.1}$^\circ$ & 27.5 & 60.6 & 88.4 & 92.3 & 93.8 &  0.20/{0.08} \\
			& (ScaN + ArkitS + ScaNP + 3DM) & \textbf{69.4} &  \textbf{86.9} &  \textbf{94.7} &  \textbf{95.8} &  \textbf{96.1} &  \textbf{6.7}$^\circ$/\textbf{2.1}$^\circ$ & \textbf{36.2} & \textbf{71.8} & \textbf{92.5} & \textbf{95.4} & \textbf{96.1} &  \textbf{0.15}/\textbf{0.07} \\
			\midrule[1pt]
			\multirow{5}{*}{\textbf{FUSER-DF}} &(ScaN) \textit{w/o} 2D Attention Prior & 33.3 & 46.2 & 58.8 & 69.2 & 73.6 & 32.6$^\circ$/6.0$^\circ$ & 18.3 & 34.0 & 51.0 & 62.1 & 69.8 &  0.67/0.24  \\
			& (ScaN) & 51.0 & 67.1 & 79.7 & 84.9 & 85.4 & 21.4$^\circ$/2.9$^\circ$ & 28.1 & 48.3 & 72.3 & 81.5 & 84.6 &  0.42/0.11  \\
			& (ScaN + ArkitS) &  55.2 & 77.2 & 91.9 & 93.1 & 93.6 & 10.6$^\circ$/2.4$^\circ$ & 29.2 & 59.4 & 86.0 & 91.9 & 93.7 &  {0.23/0.08} \\
			& (ScaN + ArkitS + ScaNP) & {66.8} & {84.9} & {91.3} & {93.0} & {93.4} &  {9.2$^\circ$/2.1$^\circ$} & 39.0 & {68.6} & {89.3} & {92.2} & {93.6} &  0.20/\textbf{0.06} \\
			& (ScaN + ArkitS + ScaNP + 3DM) &  \textbf{72.0} & \textbf{89.7} & \textbf{94.5} & \textbf{95.2} & \textbf{95.5} &  \textbf{7.1}$^\circ$/\textbf{2.0}$^\circ$ & \textbf{43.2} & \textbf{75.5} & \textbf{92.8} & \textbf{94.7} & \textbf{95.6} &  \textbf{0.15}/\textbf{0.06}\\
			\bottomrule[1.8pt]
	\end{tabular}}
	\label{abla1}
\end{table*}

\section{Experiments}
\subsection{Experimental Settings}
\noindent\textbf{Implementation Details.} 
We train {FUSER} with sequence lengths randomly sampled from 2 to 50 to accommodate varying scan set sizes. The loss weights $\gamma_t$ and $\gamma_p$ in loss~\ref{loss2} are both 0.1, and the Huber threshold $\beta$ in loss~\ref{pointloss} is 0.06.
For {FUSER-DF}, we follow~\cite{jiang2023se} with $T{=}200$ diffusion steps during training and 10 denoising steps at inference, setting the perturbation weight $\gamma{=}0.1$ in Eq.~\ref{diff2}. 
We train on four large-scale indoor datasets, 3DMatch~\cite{zeng20173dmatch}, ScanNet~\cite{dai2017scannet}, ScanNet++~\cite{yeshwanth2023scannet++}, and ArkitScenes~\cite{baruch2021arkitscenes}, for 2 epochs using the Adam optimizer (learning rate $5{\times}10^{-5}$, weight decay 0.05) with a linear decay schedule down to $1{\times}10^{-8}$. 
Training uses 8 NVIDIA L20 GPUs (45 GB each), and evaluation uses a single GPU. FUSER has $\sim$0.6B parameters. 
% The model is trained on 8 NVIDIA L20 GPUs (45 GB memory each) a
% All experiments are executed in PyTorch on 8 NVIDIA L20 GPUs (45 GB memory each) a for training and . FUSER has $\sim$592M parameters.

\noindent\textbf{Metrics.} We evaluate registration accuracy using rotation and translation errors (RE/TE) on ScanNet and Registration Recall (RR) on 3DMatch and ArkitScenes, as in~\cite{wang2023robust,yew2021learning,gojcic2020learning,choy2020deep}.
More details are provided in  Appendix A. 

\subsection{Comparison with Existing Methods}
\noindent\textbf{Evaluation on ScanNet.}
We first  evaluate {FUSER} and {FUSER-DF} on ScanNet~\cite{dai2017scannet}, a well-known indoor RGB-D dataset.
Following~\cite{wang2023robust}, we use total 32 sequences with 30 scans each (20 frames apart) for evaluation. 
Baselines include EIGSE3~\cite{arrigoni2016spectral}, L1-IRLS~\cite{chatterjee2017robust}, RotAvg~\cite{govindu2004lie}, LMVR~\cite{gojcic2020learning}, LITS~\cite{yew2021learning}, HARA~\cite{lee2022hara}, SGHR~\cite{wang2023robust}, MDGD~\cite{li2024matching}, and IncreMVR~\cite{cheng2024incremental}.
All methods use YOHO~\cite{wang2022you} descriptors for pairwise registration except LMVR, which performs end-to-end alignment and synchronization.
As~\cite{jin2024multiway} has not released code, it is excluded.  
We follow~\cite{gojcic2020learning,wang2023robust} and consider three pose graphs:
\textbf{(i)} \textit{Full}: exhaustive pairwise registration;
\textbf{(ii)} \textit{Pruned}: filtering pairs by a median point distance threshold (0.05m); and
\textbf{(iii)} \textit{Sparse}: adaptively selecting pairs based on overlap estimation.

\begin{figure}[t]
	\centering
	\includegraphics[width=\columnwidth]{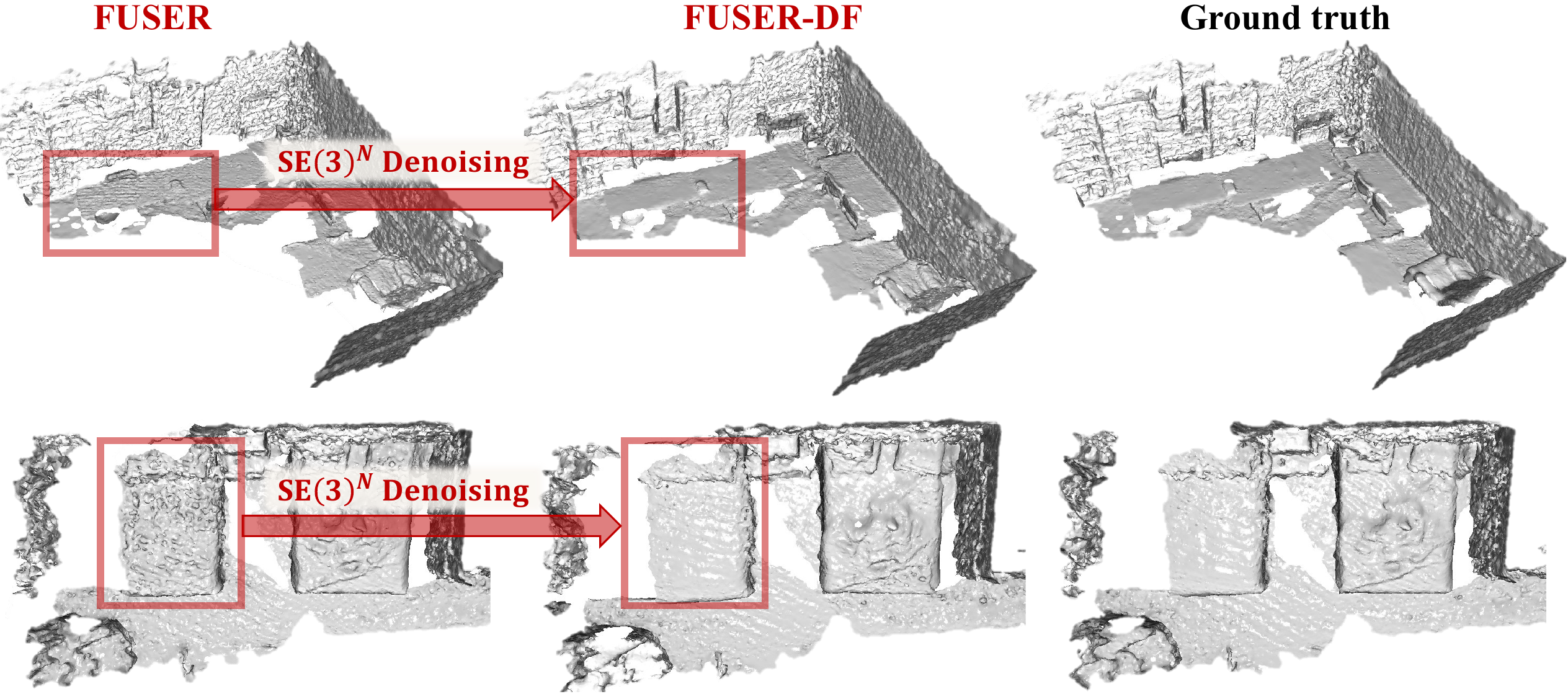}
	\vspace{-6mm}
	\caption{ SE(3)$^N$ diffusion refinement in FUSER-DF visually refines FUSER’s pose estimation, yielding smoother surfaces.}
	\label{visse3}
	\vspace{-6mm}
\end{figure}

\begin{figure}[t]
	\centering
	\includegraphics[width=\columnwidth]{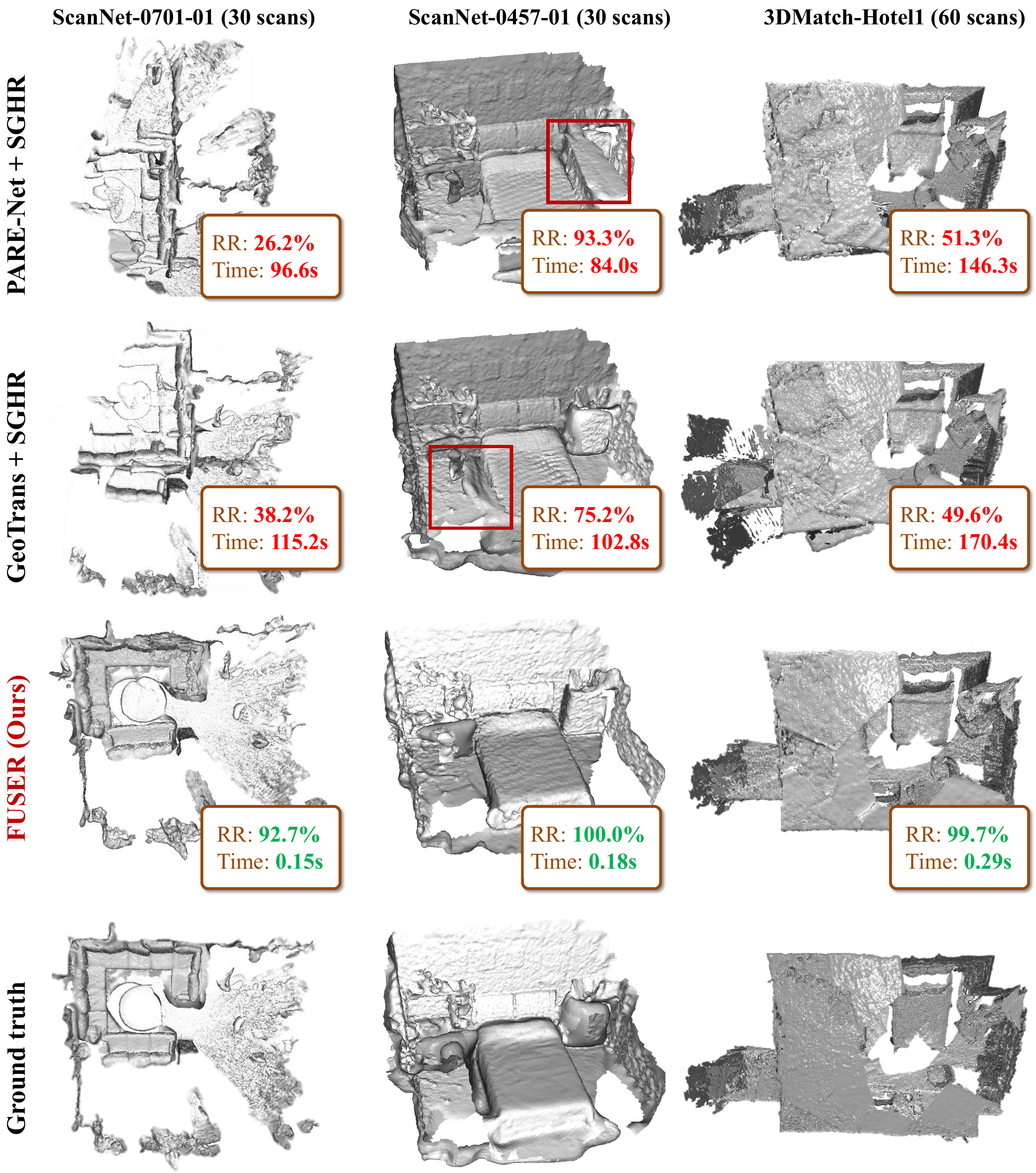}
	\vspace{-6mm}
	\caption{ Qualitative comparison: FUSER surpasses SOTA GeoTrans~\cite{qin2022geometric} and PARENet~\cite{yao2024pare} descriptors with SGHR pose graph~\cite{wang2023robust}, achieving much higher accuracy and efficiency.}
	\label{viscomp10}
	\vspace{-3mm}
\end{figure}

As shown in Table~\ref{scannetcompa}, both {FUSER} and {FUSER-DF} consistently surpass all baselines across nearly all metrics.
Compared with the strongest prior method~\cite{li2024matching}, {FUSER} significantly reduces mean translation and rotation errors (0.37\textit{m}$\rightarrow$0.15\textit{m}, 17.4$^\circ$$\rightarrow$6.7$^\circ$), highlighting the advantage of our feed-forward design that avoids cumulative pose noise in traditional pairwise pipelines.
Further, {FUSER-DF} refines the multiview alignment of {FUSER} via the proposed SE(3)$^N$ diffusion process, yielding notable gains under strict thresholds (e.g., Rot@3°: 69.4 → 72.0) while maintaining high overall accuracy (see Fig.~\ref{visse3} and Fig.~\ref{viscomp10}).

\noindent\textbf{Evaluation on 3DMatch.}
We further evaluate our framework on the 3DMatch dataset~\cite{zeng20173dmatch}, a widely used indoor RGB-D dataset. Following the eight testing scenes in~\cite{wang2023robust}, we conduct multiview alignment under a more realistic and challenging setup. 
Unlike prior 3DMatch benchmark that fuses 50 consecutive frames into GT-aligned fragments via TSDF integration (an operation that effectively denoises, alleviates occlusions, and increases geometric overlap), we construct each sequence directly from single frame to better reflect real-world deployment with limited overlap and sensor noise. Each sequence contains 60 frames with a 20-frame interval (consistent with ScanNet). 
We combine five SOTA deep descriptors, FCGF~\cite{choy2019fully}, Predator~\cite{huang2021predator}, YOHO~\cite{wang2022you}, GeoTrans~\cite{qin2022geometric}, and PARENet~\cite{yao2024pare}, with two pose-graph types, \textit{full} and SGHR-based \textit{sparse}, for comparison. 
As shown in Table~\ref{3dmatchcompa}, both {FUSER} and {FUSER-DF} achieve superior accuracy across all metrics. %Our feed-forward design removes redundant pairwise registrations, reducing runtime from minutes to under one second (155.4\textit{s}$\rightarrow$0.31\textit{s}), while diffusion refinement remains efficient (2.91\textit{s}), demonstrating high accuracy, speed, and scalability. 

\noindent\textbf{Evaluation on ArKitScenes.} We finally perform evaluation on ArKitScenes~\cite{baruch2021arkitscenes}, an indoor RGB-D dataset captured with LiDAR-equipped iPhones. 
We randomly select 15 sequences, each with 200 scans sampled every three frames. 
For a fair evaluation, we select the two best-performing baselines on 3DMatch for comparison.
% Two best-performed baselines in 3DMatch are selected for comparisons. 
Table~\ref{ark} demonstrates that our method exhibits strong precision, robustness, and generalizability under large-scale scan settings.

\begin{table}[t]
	\centering
	\caption{Runtime (s) on \textbf{3DMatch}~\cite{zeng20173dmatch} and \textbf{ArKitScenes}~\cite{baruch2021arkitscenes}.}
	\vspace{-3mm}
	\setlength{\tabcolsep}{3pt}
	\resizebox{1\columnwidth}{!}{
		\begin{tabular}{c lccccccc}
			\toprule[1.8pt]
			& & FCGF & Pred. & YOHO & GeoT. & PARE. & \textcolor[RGB]{166,35,35}{\textbf{FUSER}} & \textcolor[RGB]{166,35,35}{\textbf{FUSER-DF}} \\
			\midrule[1pt]
			
			\multirow{2}{*}{\rotatebox{90}{\textbf{3DM}}}
			& \textbf{Full} & 517.4 & 690.7 & 762.9 & 495.4 & 384.0 & \multirow{2}{*}{\textbf{0.31}}  & \multirow{2}{*}{\textbf{2.91}} \\
			& \textbf{SGHR} & 292.5 & 316.6 & 517.5 & 180.5 & 155.4 & & \\
			
			\midrule[1pt]
			
			\multirow{2}{*}{\rotatebox{90}{\textbf{ArK}}}
			& \textbf{Full} & -- & -- &  -- & 2454.6 & 1831.3 & \multirow{2}{*}{\textbf{0.61}} & \multirow{2}{*}{\textbf{6.50}} \\
			& \textbf{SGHR} & -- & -- & -- & 714.8 & 583  &  & \\
			\bottomrule[1.8pt]
	\end{tabular}}
	\label{tab:speed}
	\vspace{-6mm}
\end{table}

\vspace{-2mm}
\subsection{Ablation Studies and Analysis}
\vspace{-1mm}
\noindent\textbf{Scaling with Data Volume.}
Table~\ref{abla1} shows that with increasing training data, our model yields consistent accuracy gains, indicating a favorable scaling trend and strong potential for large-scale 3D foundation training.

\noindent\textbf{Effectiveness of 2D Attention Prior.}
We evaluate the impact of the 2D Attention Prior from pretrained 2D foundation model, $\pi^3$~\cite{wang2025pi}. 
Table~\ref{abla1} shows that integrating this prior significantly enhances both rotation and translation precision.
This demonstrates that injecting 2D attention priors via parameter initialization substantially improves geometric reasoning, leading to more robust multiview alignment. 

% Comparing the ``w/o 2D Attention Prior'' and full variants on ScanNet (\textit{ScaN}), clear improvements are observed in both rotation and translation metrics. These gains show that injecting 2D attention priors via parameter initialization can significantly enhance  geometric reasoning and promotes more robust multiview alignment.

\noindent\textbf{Runtime and Memory Analysis.}
Table~\ref{tab:speed} shows prior two-stage pipelines that rely on redundant pairwise registrations are highly time-consuming, requiring hundreds of seconds per sequence on {3DMatch} (60 scans) and even thousands of seconds on {ArkitScenes} (200 scans).
By contrast, our feed-forward paradigm, {FUSER}, and its diffusion variant, {FUSER-DF}, remove the redundant pairwise optimization, enabling efficient second-scale inference while maintaining superior accuracy, as evidenced in Tables~\ref{3dmatchcompa} and~\ref{ark}. 
Moreover, thanks to the highly compact superpoint representation (Sec.~\ref{age}) and efficient FlashAttention~\cite{dao2022flashattention,dao2023flashattention2}, FUSER and FUSER-DF consume only 2.83G and 5.09G of GPU memory, respectively, for the longest ArkitScenes sequences (200 scans), highlighting its practical efficiency. 

% Moreover, thanks to the highly compact superpoint representation (Sec.~\ref{age}) and the efficient attention computation of FlashAttention~\cite{dao2022flashattention,dao2023flashattention2}, FUSER and FUSER-DF consumes only 2.83G and 5.09G of GPU memory per ArkitScenes scene with longest 200 scans, highlighting its practical efficiency. 
%Thanks to the effecient attention computation via FlashAttention, the GPU Consumption per scene under ArkitScenes benchmark with longest sequence of 200 scans only requires 2.83G GPU memory.  

\vspace{-2mm}
\section{Conclusion}
\vspace{-2mm}
\label{sec:conclusion}
We presented FUSER, the first feed-forward multiview registration transformer that unifies all scans within a compact latent space for direct global pose prediction, eliminating redundant pairwise estimation.
By integrating absolute geometric encoding, 2D attention priors, and geometric alternating attention, FUSER achieves holistic cross-scan reasoning and surpasses previous methods by a large margin, while cutting inference time from minutes to the second level.
We further introduced FUSER-DF, an SE(3)$^N$ diffusion refinement framework that formulates multiview registration as a denoising process for fine-grained pose correction.
Both of them marked a paradigm shift from two-stage pipeline to unified feed-forward paradigm, advancing 3D foundation modeling for multiview registration. 

% \noindent\textbf{Limitations and Future Work.} Our experiments show consistent performance gains with increasing data volume, and we plan to further explore how FUSER scales with model capacity.
% Moreover, while our current work targets indoor scenes, extending our method to outdoor scenes will be an exciting future direction. 

\section*{Acknowledgments}
This research is supported by the MOE AcRF Tier 1 Grant of Singapore (RG107/25), by the RIE2025 Industry Alignment Fund – Industry Collaboration Projects (IAF-ICP) (Award I2301E0026), administered by A*STAR, as well as supported by Alibaba Group and NTU Singapore through Alibaba-NTU Global e-Sustainability CorpLab (ANGEL).

%\vspace{-2mm}
%\section{Limitations and Future Work}
%\vspace{-2mm}
%\label{limite}
%Our experiments show consistent performance gains with increasing data volume, and we plan to further explore how FUSER scales with model capacity.
%Moreover, while our current work targets indoor scenes, extending our method to outdoor scenes will be an exciting future direction. 
%\input{2_formatting}
%\section{Final copy}
%
%You must include your signed IEEE copyright release form when you submit your finished paper.
%We MUST have this form before your paper can be published in the proceedings.
%
%Please direct any questions to the production editor in charge of these proceedings at the IEEE Computer Society Press:
%\url{https://www.computer.org/about/contact}.
{
    \small
    \bibliographystyle{ieeenat_fullname}
    \bibliography{main}
}

% \clearpage
% \onecolumn
% \setcounter{page}{1}

% \setcounter{section}{0}
% \renewcommand{\thesection}{\Alph{section}}
% \renewcommand{\thesubsection}{\thesection.\arabic{subsection}}

% \maketitlesupplementary

\clearpage
\onecolumn
\setcounter{page}{1}

% Section A, B, C ...
\setcounter{section}{0}
\renewcommand{\thesection}{\Alph{section}}
\renewcommand{\thesubsection}{\thesection.\arabic{subsection}}

\begin{center}
{\Large \bf Supplementary Material for ``{FUSER}:  {F}eed-Forward M{U}ltiview 3D Registration Transformer and  {SE}(3)$^N$ Diffusion {R}efinement''}\\[1em]
\end{center}

\section{Evaluation Metrics}
We evaluate multiview registration accuracy by comparing predicted relative poses ${\hat{\mathbf{R}}_{ij}, \hat{\mathbf{t}}_{ij}}$ with ground truth ${{\mathbf{R}}_{ij}, {\mathbf{t}}_{ij}}$.
For ScanNet~\cite{dai2017scannet}, following~\cite{wang2023robust,yew2021learning,gojcic2020learning}, we report the empirical cumulative distribution function (ECDF) of rotation/translation errors:
\begin{equation}\small
	\begin{split}
		\operatorname{RE}_{ij} = \operatorname{arccos}\frac{\operatorname{Tr}\left(\hat{\mathbf{R}}_{ij}^\top\mathbf{R}_{ij}\right)-1}{2}, \ \ \operatorname{TE}_{ij} = \|\hat{\mathbf{t}}_{ij}-{\mathbf{t}}_{ij}\|_2
	\end{split}
\end{equation}
For 3DMatch~\cite{zeng20173dmatch}, following~\cite{choy2020deep}, we report \textit{Registration Recall} (RR), the percentage of successful registrations within 15° and 0.3 m thresholds, together with average rotation (\textit{RE}) and translation (\textit{TE}) errors. 
We also list the number of pairwise matchs ({\textit{\#Pair}}) executed by each method. 

\section{Variational Lower Bound Derivation for Prior-aware SE(3)$^N$ Diffusion Refinement Model}

% \subsection{Initial VLB with Jensen's Inequality}
The objective is to find a tractable lower bound on the marginal log-likelihood of the ground-truth transformations $\mathbf{T}^0_{1:N}$ given the data $\mathcal{S}=\{\mathbf{S}_1, \mathbf{S}_2, ..., \mathbf{S}_N\}$ and the prior transformations $\hat{\mathbf{T}}_{1:N}=(\hat{\mathbf{T}}_1, \hat{\mathbf{T}}_2, ..., \hat{\mathbf{T}}_N)$ predicted by the FUSER. We introduce the set of latent transformation variables $\mathbf{T}_{1:N}^{1:T}=(\mathbf{T}_{1:N}^1, \mathbf{T}_{1:N}^2, \ldots, \mathbf{T}_{1:N}^T)$ and apply the Jensen's Inequality (I):

\begin{equation}\label{eq:vlb_start}
	\begin{split}
		& \ln p_\theta(\mathbf{T}^0_{1:N}\mid\mathcal{S}, \hat{\mathbf{T}}_{1:N}) \\
		&= \ln\int_{\mathbf{T}_{1:N}^{1:T}}p_\theta(\mathbf{T}_{1:N}^{0:T}\mid \mathcal{S}, \hat{\mathbf{T}}_{1:N})d_{\mathbf{T}_{1:N}^{1:T}} \\
		&= \ln\int_{\mathbf{T}_{1:N}^{1:T}}q(\mathbf{T}_{1:N}^{1:T}\mid \mathbf{T}_{1:N}^0, \hat{\mathbf{T}}_{1:N}) \frac{p_\theta(\mathbf{T}_{1:N}^{0:T}\mid \mathcal{S}, \hat{\mathbf{T}}_{1:N})}{q(\mathbf{T}_{1:N}^{1:T}\mid \mathbf{T}_{1:N}^0, \hat{\mathbf{T}}_{1:N})}d_{\mathbf{T}_{1:N}^{1:T}} \\
		&= \ln\mathbb{E}_{\mathbf{T}_{1:N}^{1:T}\sim q}\left[\frac{p_\theta(\mathbf{T}_{1:N}^{0:T}\mid \mathcal{S}, \hat{\mathbf{T}}_{1:N})}{q(\mathbf{T}_{1:N}^{1:T}\mid \mathbf{T}_{1:N}^0, \hat{\mathbf{T}}_{1:N})}\right] \\
		&\stackrel{\text{(I)}}{\geq} \mathbb{E}_{\mathbf{T}_{1:N}^{1:T}\sim q}\left[\ln\frac{p_\theta(\mathbf{T}_{1:N}^{0:T}\mid \mathcal{S}, \hat{\mathbf{T}}_{1:N})}{q(\mathbf{T}_{1:N}^{1:T}\mid \mathbf{T}_{1:N}^0, \hat{\mathbf{T}}_{1:N})}\right]
	\end{split}
\end{equation}

%\subsection{Chain Rule Factorization of Prior and Posterior}
Based on the defined conditional dependencies in the diffusion process, the prior distribution $p_\theta$ and the posterior distribution $q$ can be factorized as follows:
\begin{equation}
\begin{split}
p_\theta(\mathbf{T}_{1:N}^{0:T}\mid \mathcal{S}, \hat{\mathbf{T}}_{1:N}) = p(\mathbf{T}_{1:N}^T\mid \hat{\mathbf{T}}_{1:N}) \cdot p_\theta(\mathbf{T}_{1:N}^{0:T-1}\mid \mathcal{S}_T, \hat{\mathbf{T}}_{1:N}) = p(\mathbf{T}_{1:N}^T\mid \hat{\mathbf{T}}_{1:N}) \cdot \prod_{t=1}^{T} p_\theta(\mathbf{T}_{1:N}^{t-1}\mid \mathcal{S}_t, \hat{\mathbf{T}}_{1:N})
\end{split}
\end{equation}

\begin{equation}
\begin{split}
q(\mathbf{T}_{1:N}^{1:T}\mid \mathbf{T}_{1:N}^0, \hat{\mathbf{T}}_{1:N}) = \prod_{t=1}^{T} q(\mathbf{T}_{1:N}^t\mid \mathbf{T}_{1:N}^{t-1}, \hat{\mathbf{T}}_{1:N})
\end{split}
\end{equation}
where $\mathcal{S}_t$ represents the transformed point-cloud scans at timestep $t$. By substituting the factorized distributions into Eq.~\ref{eq:vlb_start}, we can achieve:
\begin{equation}\label{eq:vlb_factorized}
	\begin{split}
		& \ln p_\theta(\mathbf{T}^0_{1:N}\mid\mathcal{S}, \hat{\mathbf{T}}_{1:N}) \\
		\geq & \mathbb{E}_{q}\left[\ln\frac{p(\mathbf{T}_{1:N}^T\mid \hat{\mathbf{T}}_{1:N}) \prod_{t=1}^{T} p_\theta(\mathbf{T}_{1:N}^{t-1}\mid \mathcal{S}_t, \hat{\mathbf{T}}_{1:N})}{\prod_{t=1}^{T} q(\mathbf{T}_{1:N}^t\mid \mathbf{T}_{1:N}^{t-1}, \hat{\mathbf{T}}_{1:N})}\right] \\
		= & \mathbb{E}_{q}\left[\ln p(\mathbf{T}_{1:N}^T\mid \hat{\mathbf{T}}_{1:N}) + \sum^{T}_{t=1}\ln\frac{p_\theta(\mathbf{T}_{1:N}^{t-1}\mid \mathcal{S}_t, \hat{\mathbf{T}}_{1:N})}{q(\mathbf{T}_{1:N}^t\mid \mathbf{T}_{1:N}^{t-1}, \hat{\mathbf{T}}_{1:N})}\right]
	\end{split}
\end{equation}

We use Bayes' formula to deform the forward transition $q(\mathbf{T}_{1:N}^t\mid \mathbf{T}_{1:N}^{t-1}, \hat{\mathbf{T}}_{1:N})$:
$$
q(\mathbf{T}_{1:N}^t\mid \mathbf{T}_{1:N}^{t-1}, \hat{\mathbf{T}}_{1:N}) = \frac{q(\mathbf{T}_{1:N}^{t-1}\mid \mathbf{T}_{1:N}^t, \mathbf{T}_{1:N}^0, \hat{\mathbf{T}}_{1:N}) q(\mathbf{T}_{1:N}^t\mid \mathbf{T}_{1:N}^0, \hat{\mathbf{T}}_{1:N})}{q(\mathbf{T}_{1:N}^{t-1}\mid \mathbf{T}_{1:N}^0, \hat{\mathbf{T}}_{1:N})}
 $$
Substituting this into Eq.~\ref{eq:vlb_factorized} yields:

\begin{equation}\label{eq:vlb_bayes}
	\begin{split}
		& \ln p_\theta(\mathbf{T}^0_{1:N}\mid\mathcal{S}, \hat{\mathbf{T}}_{1:N}) \\
		\geq & \mathbb{E}_{q}\left[\ln p(\mathbf{T}_{1:N}^T\mid \hat{\mathbf{T}}_{1:N}) + \sum^{T}_{t=1}\ln\frac{p_\theta(\mathbf{T}_{1:N}^{t-1}\mid \mathcal{S}_t, \hat{\mathbf{T}}_{1:N})}{q(\mathbf{T}_{1:N}^t\mid \mathbf{T}_{1:N}^{t-1}, \hat{\mathbf{T}}_{1:N})}\right] \\
		= & \mathbb{E}_{q}\left[\ln p(\mathbf{T}_{1:N}^T\mid \hat{\mathbf{T}}_{1:N}) + \sum^{T}_{t=1}\ln\frac{p_\theta(\mathbf{T}_{1:N}^{t-1}\mid \mathcal{S}_t, \hat{\mathbf{T}}_{1:N})q(\mathbf{T}_{1:N}^{t-1}\mid \mathbf{T}_{1:N}^0, \hat{\mathbf{T}}_{1:N})}{q(\mathbf{T}_{1:N}^{t-1}\mid \mathbf{T}_{1:N}^t, \mathbf{T}_{1:N}^0, \hat{\mathbf{T}}_{1:N}) q(\mathbf{T}_{1:N}^t\mid \mathbf{T}_{1:N}^0, \hat{\mathbf{T}}_{1:N})}\right] \\
    = & \mathbb{E}_{q}\left[\ln p(\mathbf{T}_{1:N}^T\mid \hat{\mathbf{T}}_{1:N}) + \sum^{T}_{t=2}\ln\frac{p_\theta(\mathbf{T}_{1:N}^{t-1}\mid \mathcal{S}_t, \hat{\mathbf{T}}_{1:N})q(\mathbf{T}_{1:N}^{t-1}\mid \mathbf{T}_{1:N}^0, \hat{\mathbf{T}}_{1:N})}{q(\mathbf{T}_{1:N}^{t-1}\mid \mathbf{T}_{1:N}^t, \mathbf{T}_{1:N}^0, \hat{\mathbf{T}}_{1:N}) q(\mathbf{T}_{1:N}^t\mid \mathbf{T}_{1:N}^0, \hat{\mathbf{T}}_{1:N})}+ \ln\frac{p_\theta(\mathbf{T}_{1:N}^{0}\mid \mathcal{S}_1, \hat{\mathbf{T}}_{1:N})\cdot 1}{1\cdot q(\mathbf{T}_{1:N}^1\mid \mathbf{T}_{1:N}^0, \hat{\mathbf{T}}_{1:N})}\right] \\
    = & \mathbb{E}_{q}\Bigg[\ln{p_\theta(\mathbf{T}_{1:N}^0\mid \mathcal{S}_1, \hat{\mathbf{T}}_{1:N})} -\operatorname{D_{KL}}(q(\mathbf{T}_{1:N}^T\mid \mathbf{T}_{1:N}^0, \hat{\mathbf{T}}_{1:N}) || p(\mathbf{T}_{1:N}^T\mid \hat{\mathbf{T}}_{1:N})) \\
		& - \sum^{T}_{t=2}\operatorname{D_{KL}}(q(\mathbf{T}_{1:N}^{t-1}\mid \mathbf{T}_{1:N}^t, \mathbf{T}_{1:N}^0, \hat{\mathbf{T}}_{1:N}) || {p_\theta(\mathbf{T}_{1:N}^{t-1}\mid\mathcal{S}_t, \hat{\mathbf{T}}_{1:N})})
		\Bigg]
		\end{split}
\end{equation}
This completes the derivation.

\section{Model Architecture of Absolute Geometric Encoder}
\begin{figure}[ht]
		\centering
		\includegraphics[width=0.7\textwidth]{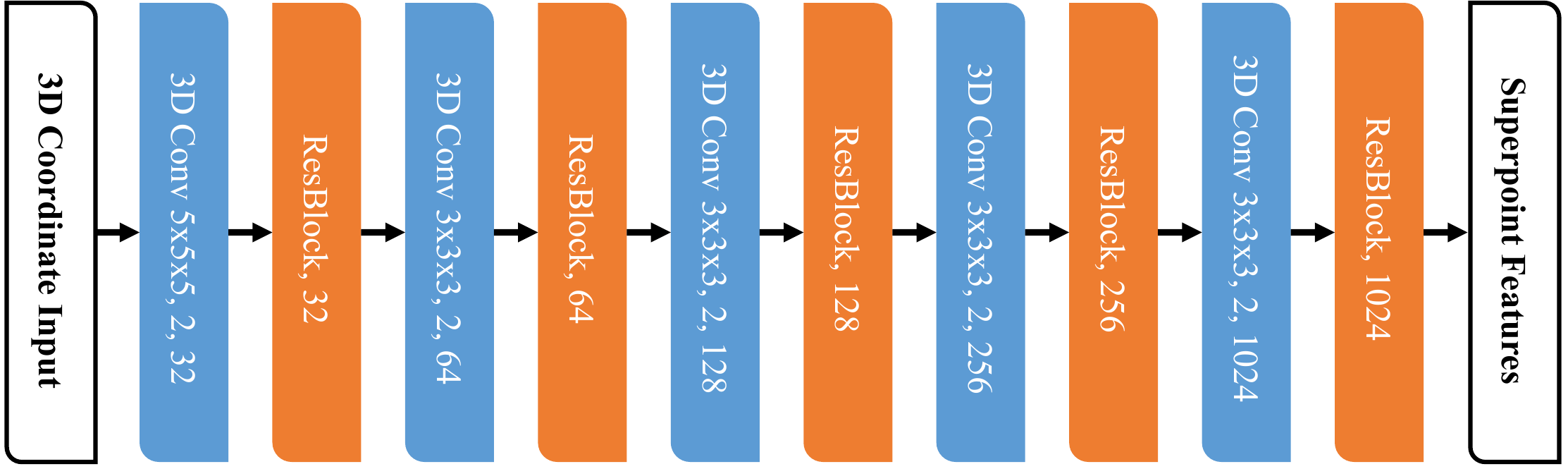}
		\caption{Network architecture of absolute geometric encoder.} 
		\label{graph2}
	\end{figure}

We present the detailed architecture of the absolute geometric encoder in Fig.~\ref{graph2}. 

\section{In-the-Wild Evaluation}
\begin{figure}[ht]
    \centering
    \includegraphics[width=1\textwidth]{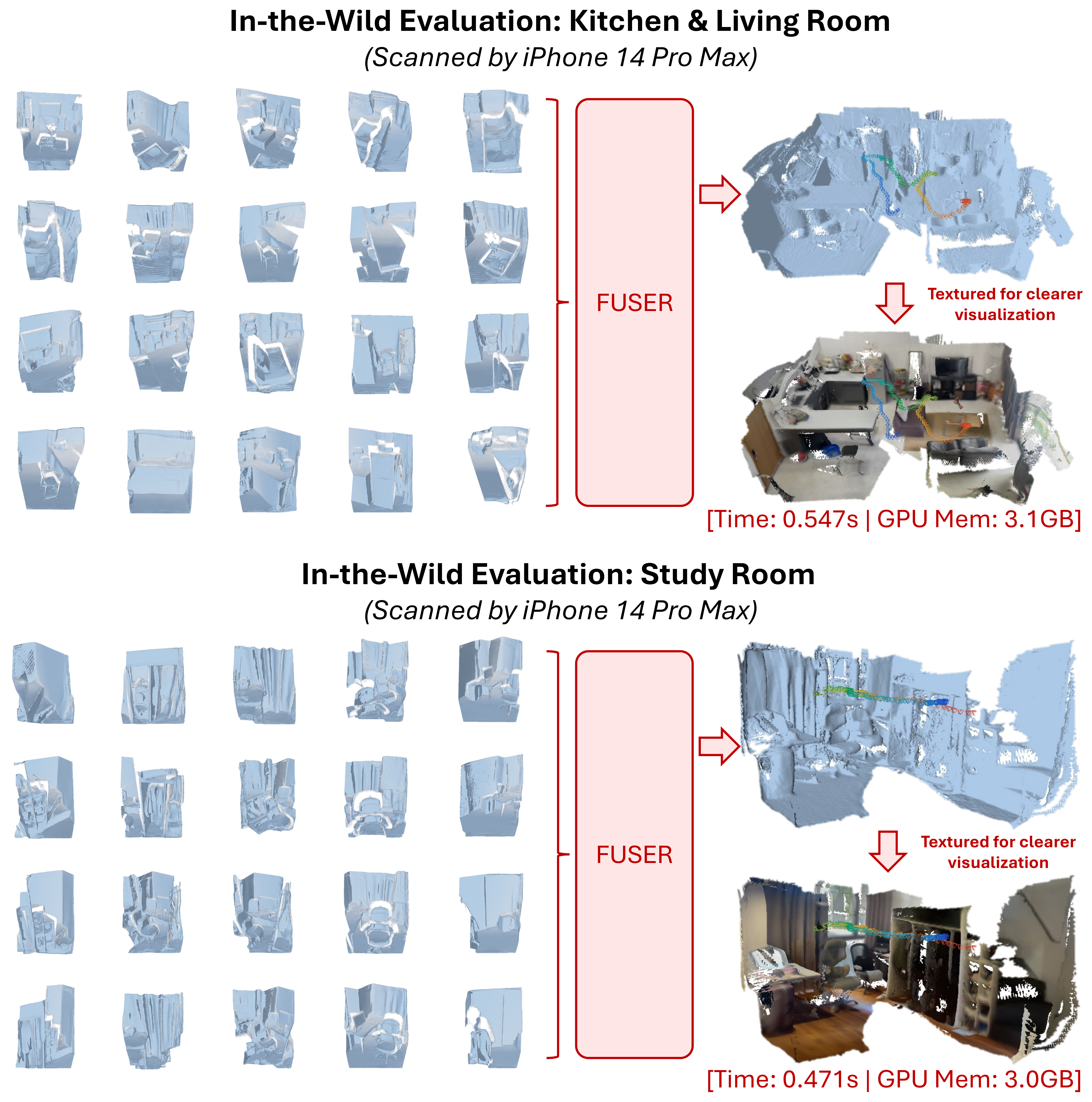}
    \caption{In-the-wild evaluation of FUSER on real-world multiview 3D sequences acquired using an iPhone 14 Pro Max.} 
    \label{inthewild}
	\end{figure}

In addition to benchmark evaluations, we further evaluate FUSER on in-the-wild point-cloud sequences acquired with an iPhone 14 Pro Max in the authors’ kitchen/living room and study room. Despite challenges such as point sparsity, geometric distortion, and measurement noise, FUSER achieves high-quality feed-forward multiview registration, as shown in Fig.~\ref{inthewild}, demonstrating strong robustness and generalizability in real-world scenarios. We note that textures are added solely for clearer visualization; the actual input to FUSER consists only of point-cloud data. 

% \begin{equation}\small \label{vlb}
% 	\begin{split}
% 		\setlength{\abovedisplayskip}{2pt}
% 		\setlength{\belowdisplayskip}{2pt}
% 		& \mathbb{E}_{\mathcal{S}, \mathbf{T}^0_{1:N}\sim p_{data}}\left[\ln p_\theta(\mathbf{T}^0_{1:N}\mid\mathcal{S}, \hat{\mathbf{T}}_{1:N}) \right] \\
% 		\geq & \mathbb{E}\left[\ln\frac{p_\theta(\mathbf{T}^{0:T}_{1:N}\mid \mathcal{S}, \hat{\mathbf{T}}_{1:N})}{q(\mathbf{T}_{1:N}^{1:T}\mid \mathbf{T}_{1:N}^0, \hat{\mathbf{T}}_{1:N})}\right]=\underbrace{\mathbb{E}\left[\ln{p_\theta(\mathbf{T}_{1:N}^0\mid\mathcal{S}, \hat{\mathbf{T}}_{1:N})}\right]}_{\text{Residual term}} \\
% 		& - \underbrace{\mathbb{E}\left[\operatorname{{KL}}(q(\mathbf{T}_{1:N}^T\mid \mathbf{T}_{1:N}^0, \hat{\mathbf{T}}_{1:N}) || p(\mathbf{T}_{1:N}^T\mid\hat{\mathbf{T}}_{1:N}))\right]}_{\text{Prior matching term}} \\
% 		- & \mathbb{E}\Big[\sum^{T}_{t=2}\underbrace{\operatorname{{KL}}(q(\mathbf{T}_{1:N}^{t-1}\mid \mathbf{T}_{1:N}^t, \mathbf{T}_{1:N}^0, \hat{\mathbf{T}}_{1:N}) || {p_\theta(\mathbf{T}_{1:N}^{t-1}\mid\mathcal{S}_t, \hat{\mathbf{T}}_{1:N})})}_{\text{Prior-aware Denoising matching term}} \Big],
% 	\end{split}
% \end{equation}

% {
%     \small
%     \bibliographystyle{ieeenat_fullname}
%     \bibliography{main}
% }

\end{document}